%% file: acl_latex.tex
\pdfoutput=1
\documentclass[11pt]{article}

\usepackage{acl}

\usepackage{times}
\usepackage{latexsym}

\usepackage[T1]{fontenc}
\usepackage[utf8]{inputenc}

\usepackage{microtype}

\usepackage{inconsolata}

\usepackage{graphicx}

\usepackage{inconsolata}
\usepackage{graphicx}

\usepackage{enumitem}
\usepackage{multirow}
\usepackage{booktabs}
\usepackage{algorithm}
\usepackage{algorithmic}
\usepackage{tikz}
\usepackage{pgfplots}
\usepackage{graphicx}
\usepackage{subcaption}
\usepackage{pgf-pie}
\usepackage{enumitem}
\usepackage{pifont}
\usepackage{amsmath}
\usepackage{supertabular}
\numberwithin{equation}{section}
\usepackage{dsfont}

\pgfplotsset{compat=1.18}
\usepackage{bm}
\usepackage{xspace}
\usepackage{tcolorbox}
\usepackage{bbding}
\usepackage{wasysym}
\usepackage{amssymb}
\usepackage{fontawesome5}
\usepackage{arydshln}  
\usepackage{dashrule}
\usepgfplotslibrary{dateplot}
\usepackage{mathtools}
\usepackage{nicematrix}

\newcommand{\ours}{\textsc{RoT}\xspace}
\definecolor{cblue}{rgb}{0.58, 0.86, 0.91}
\definecolor{sblue}{rgb}{0.859, 0.906, 0.961}
\definecolor{dblue}{HTML}{3E95BC}
\definecolor{cred}{rgb}{0.98, 0.45, 0.31}
\definecolor{dred}{HTML}{5b0f00}
\definecolor{cwhite}{rgb}{0.96, 0.95, 0.91}
\definecolor{cyellow}{rgb}{0.90, 0.87, 0.74}
\definecolor{syellow}{rgb}{0.914, 0.890, 0.808}
\definecolor{cgreen}{rgb}{0.255, 0.525, 0.373}
\definecolor{sgreen}{rgb}{0.878, 0.933, 0.835}
\definecolor{dgreen}{HTML}{274E13}
\definecolor{spurple}{rgb}{0.859, 0.843, 0.929}
\definecolor{dpurple}{HTML}{AFA5D7}
\definecolor{intro_green}{rgb}{0.153, 0.306, 0.075}
\definecolor{intro_blue}{rgb}{0.110, 0.271, 0.529}
\newcommand{\green}[1]{{\color{dgreen}#1}}
\newcommand{\red}[1]{{\color{dred}#1}}
\newcommand{\textintrogreen}[1]{{\color{intro_green}#1}}
\newcommand{\textintroblue}[1]{{\color{intro_blue}#1}}

\newtcolorbox{myexample}[2][]{
  colback=data_blue!40,
  colframe=data_blue,         % 设置边框颜色
  coltitle=black,
  title=\textbf{#2},
  fonttitle=\bfseries,
  #1,
}

\title{\ours: Enhancing Table Reasoning with Iterative Row-Wise Traversals}

\author{
    Xuanliang Zhang, Dingzirui Wang, Keyan Xu, Qingfu Zhu, Wanxiang Che\\ 
    \texttt{\{xuanliangzhang, dzrwang, kyxu, qfzhu, car\}@ir.hit.edu.cn}\\
    Harbin Institute of Technology 
}

\begin{document}
% \nolinenumbers
    \maketitle
    \begin{abstract}
    % 表格推理任务旨在根据表格对用户的问题进行回答，因加速了数据获取受到很多研究者关注
    The table reasoning task, crucial for efficient data acquisition, aims to answer questions based on the given table. 
    % 近些年，推理大语言模型因引入Long CoT显著提升了推理能力，在下游任务上的性能得到了提升，包括表格推理任务
    Recently, reasoning large language models (RLLMs) with Long Chain-of-Thought (Long CoT) significantly enhance reasoning capabilities, leading to brilliant performance on table reasoning.
    % 然而，获得Long CoT的能力需要高昂的训练开销
    % However, acquiring Long CoT capabilities necessitates high costs for training.
    % 并且，随着输出的变长，容易造成对表格内容的幻觉，有着低依赖性
    % Furthermore, the increased output length could lead to hallucinations of tabular contents and exhibit low reliability.
    % 然而，Long CoT遭受着high cost for training and对表格内容的幻觉，呈现出低可依赖性
    However, Long CoT suffers from high cost for training and exhibits low reliability due to table content hallucinations.
    % 为了解决以上两点限制，我们提出\ours，通过提示令模型迭代地逐行遍历表格来推理，每次迭代用于继续推理以及根据反思重新推理
    Therefore, we propose Row-of-Thought (\ours), which performs iteratively row-wise table traversal, allowing for reasoning extension and reflection-based refinement at each traversal.
    % \ours依据表格开展逐行推理，并且利用模型本身具备的反思能力，因此training-free
    Scaling reasoning length by row-wise traversal and leveraging reflection capabilities of LLMs, \ours is training-free. 
    % 并且，顺序的遍历带来了对表格内容的更多的关注，减少了幻觉
    The sequential traversal encourages greater attention to the table, thus reducing hallucinations.
    % 由于我们的方法是training-free，并且顺序的遍历带来了对表格内容的更多的关注，减少了幻觉
    % \ours is training-free and the sequential traversal encourages greater attention to tabular content, which reduces hallucinations.
    % 由于\ours依据表格逐行推理，并且self-reflection的能力本身具备
    % 我们在3个主流表格推理数据集上进行实验，在非推理模型上我们的方法平均超过在推理模型上的Long CoT 4.3%，证明了有效性，并且在WikiTQ和TableBench上取得SOTA
    Experiments show that \ours, using non-reasoning models, outperforms RLLMs by an average of $4.3\%$, and achieves state-of-the-art results on WikiTableQuestions and TableBench with comparable models, proving its effectiveness.
    % 分析实验表明相比Long CoT，我们的方法以更少的推理token达到了更好的性能，表明我们的方法更高效
    Also, \ours outperforms Long CoT with fewer reasoning tokens, indicating higher efficiency\footnote{Our code and data will be released upon acceptance.}.

    % The table reasoning task, crucial for efficient data acquisition, aims to answer questions based on the given table. Recently, reasoning large language models (RLLMs) with Long Chain-of-Thought (Long CoT) significantly enhance reasoning capabilities, leading to brilliant performance on table reasoning. However, Long CoT suffers from high cost for training and exhibits low reliability due to table content hallucinations. Therefore, we propose Row-of-Thought (RoT), which performs iteratively row-wise table traversal, allowing for reasoning extension and reflection-based refinement at each traversal. Scaling reasoning length by row-wise traversal and leveraging reflection capabilities of LLMs, RoT is training-free. The sequential traversal encourages greater attention to the table, thus reducing hallucinations. Experiments show that RoT, using non-reasoning models, outperforms RLLMs by an average of 4.3%, and achieves state-of-the-art results on WikiTableQuestions and TableBench with comparable models, proving its effectiveness. Also, RoT outperforms Long CoT with fewer reasoning tokens, indicating higher efficiency.

    \end{abstract}

    \section{Introduction}
        \input{tex/1.introduction}

    % \section{Analysis}
    %     \label{sec:analysis}
    %     \input{tex/2.analysis}
    
    \section{\ours}
        \label{sec:methodology}
        \input{tex/3.methodology}

    \section{Experiments}
        \label{sec:experiments}
        \input{tex/4.experiment}

    \section{Related Works}
        \label{sec:related}
        \input{tex/5.related}
    
    \section{Conclusion}
    % 考虑到将推理模型的Long CoT应用到表格推理任务的问题，本文关注如何不训练激发模型的表格推理能力
    Considering the limitations of Long CoT on the table reasoning task, we focus on enhancing table reasoning capabilities with low cost and high reliability. 
    % 具体来说，本文提出一个training-free的方法，通过提示模型迭代逐行遍历进行推理，直到得到最终答案
    Specifically, we propose a training-free method, \ours, which prompts the model to perform iterative row-wise traversal reasoning until the final answer is obtained. 
    % 通过对我们的方法和Long CoT进行形式化定义和推导，我们证明了Long CoT在表格推理任务上是我们的方法的子集，说明了我们方法的优越性
    % Through formal definitions and derivations, we discuss that Long CoT is a subset of \ours on the table reasoning task, highlighting the superiority of \ours.
    % 实验表明，我们的方法使用非推理模型的性能超过了Long CoT在同等规模推理模型上的性能，平均超过4.3%
    Experiments show that \ours, using non-RLLMs, outperforms Long CoT with RLLMs, achieving an average improvement of $4.3\%$, demonstrating the effectiveness of \ours.
    Additionally, \ours with RLLMs brings an average improvement of $2.4\%$ compared with Long CoT, leading to higher reliability.
    % 并且，我们的方法在WikiTableQuestions和TableBench上达到了可比规模的模型上的SOTA，证明了我们方法的有效性
    Furthermore, \ours attains SOTA performance on WikiTableQuestions and TableBench among comparable models, validating its effectiveness.
    % 分析实验说明我们的方法相比Long CoT用更少的推理token数量达到了更好的性能，并且同样激发了模型的自主反思能力
    Analysis experiments indicate that \ours with non-RLLMs achieves better performance than Long CoT with fewer reasoning tokens, showing its higher efficiency.
    % and also elicits the self-reflection ability of models.
    
    \clearpage
    
    \section*{Limitations}
    % 1. 我们没有在多轮表格问答数据集上进行实验
    (\emph{i})~We do not conduct experiments on multi-turn table question answering datasets. 
    % 我们将在未来工作中探索\ours在多轮表格问答数据集上的有效性
    We will explore the effectiveness of \ours on such datasets in future work.
    % 2. 我们只在英文数据集上进行了实验
    (\emph{ii})~Our experiments are exclusively performed on English datasets. 
    % 我们把\ours在不同语言上的实验留作未来工作
    We leave experimentation with \ours on different languages for future research.
        
    \section*{Ethics Statement}
    All models used in this paper are publicly available, and our utilization of them strictly complies with their respective licenses and terms of use.
    
    \bibliography{custom}
    
    \clearpage
    \appendix
    \label{sec:appendix}

\input{tex/6.appendix}
    
\end{document}

%% file: tex/1.introduction.tex
% 引出表格推理任务
% 表格推理是一项很重要的任务，其输入包括表格和问题，输出为根据表格对问题的回答
Table reasoning is an important task where the input consists of a question and the table, and the output is the answer based on the table \cite{jin2022table-survey,zhang2025table-survey}. 
% 其中表格由多个并列的行组成，而每行由多个cell组成，包含着大量密集的信息
Tables typically comprise multiple rows, with each row containing several information-dense cells \cite{ruan2024table-data-survey}.
% 由于自动化的表格推理可以帮助人们从表格中抽取出有效信息，提升数据获取效率，收到很多研究者关注
Automated table reasoning attracts considerable research interest due to its potential to extract valuable information from tables, thus accelerating data acquisition \cite{badaro-etal-2023-transformers-table-survey,lu2025table-survey}.

\begin{figure}[t]
    \centering
    \includegraphics[width=0.9\linewidth]{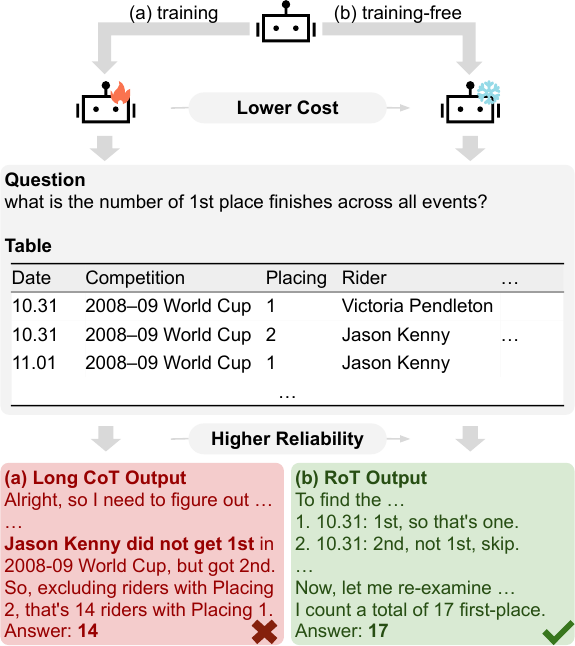}
    % \vspace{-1.5em}
    \caption{
    % \ours相比较Long CoT无需训练，有着更低的cost，并且顺序的逐行遍历缓解了幻觉问题，有着更高的可依赖性
    Compared with (a) Long CoT, (b) \ours necessitates no training, exhibits lower costs, and enhances reliability by mitigating hallucination via sequentially row-wise table traversal.
    }
    \label{fig:intro}
\end{figure}

% 推理模型及其缺点，引出我们的方法
% 近些年，推理模型凭借引入Long CoT显著提升了模型的推理能力，包括模型的表格推理能力
Recent advancements in reasoning large language models (RLLMs) have significantly enhanced reasoning capabilities utilizing Long Chain-of-Thought (Long CoT), including table reasoning capabilities \cite{li202512survey-reasoning,qian2025fino1}. 
% 因为推理模型可以进行深度的推理，自我反思，并探索多样的推理路径，相比较Short CoT浅显且直接的推理
This improvement stems from Long CoT, which sequentially scales the length of CoT, engages in self-reflection, and explores diverse reasoning paths, in contrast to the shallow and direct reasoning of Short CoT \cite{chen2025reasoning-survey,yeo2025demystifying-long-cot}.
% 然而，现在的推理模型在表格推理任务上具备以下limitation，如图1a所示
However, Long CoT exhibits two limitations in table reasoning, as illustrated in Figure~\ref{fig:intro} (a): 
% High Cost: 模型想要获得Long CoT的能力以提升表格推理能力，需要高质量的训练数据，训练成本大
(\emph{i})~\textbf{High Cost}: Achieving Long CoT capabilities for improved table reasoning capabilities necessitates high-quality data, leading to substantial training expenses \cite{qian2025fino1,jiang2024slow1}. 
% Low Reliability: 随着输出的推理链条的变长，模型容易丢失输入中的表格信息，造成幻觉
(\emph{ii})~\textbf{Low Reliability}: As the output reasoning chains lengthen, models are prone to losing relevant tabular information from the input, resulting in hallucinations of the tabular content \cite{zhang2023sirens-Hallucination-survey,liu2025mathematical-hallucination,Liu2025hallucination-long,kumar2025reliability-combining}. 

\begin{figure*}[t]
    \centering
    \includegraphics[width=0.95\linewidth]{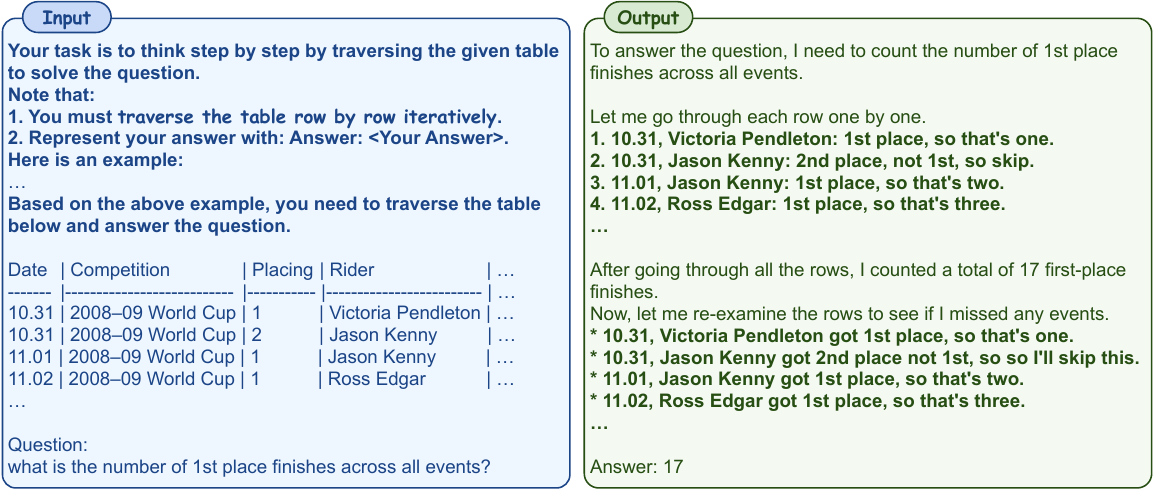}
    % \vspace{-1.5em}
    \caption{
    The overview of \ours with the input and output of the example. The instruction is highlighted with \textintroblue{\textbf{blue}} and the iterative row-wise table traversal process is highlighted with \textintrogreen{\textbf{green}}.
    }
    \label{fig:method}
\end{figure*}

% 为了解决上述限制，我们提出Row-of-Thought（\ours），提示模型迭代地逐行处理表格，来增强模型的表格推理能力，如图1b所示
Therefore, we propose \textbf{R}ow-\textbf{o}f-\textbf{T}hought (\ours), a novel method that enhances table reasoning by guiding the model to perform iteratively row-wise traversal reasoning, as illustrated in Figure~\ref{fig:intro} (b).
% 具体来说，我们的方法提示模型逐行遍历表格的同时进行推理，每次遍历结束，模型可以选择根据之前的结果继续多跳推理，也可以对之前的推理反思，并开始新的遍历以检查并纠错，如此迭代直到得到最终答案
% Specifically, \ours prompts the model to perform row-wise traversal reasoning. Following each traversal, the model can either extend its reasoning or reflect on prior steps and initiate a new traversal accordingly. This iterative process yields the final answer.
% row-wise traversal指的是模型一步只根据表格中一行的信息进行推理，并更新中间结果
Row-wise traversal refers to the reasoning process where it considers information from a single row at each step to update intermediate results.
% 迭代过程指的是，每次遍历结束，模型可以选择根据之前的结果继续多跳推理，也可以对之前的推理反思，并据此开始新的遍历
In the iterative process, after each traversal, the model can either extend its reasoning or reflect on prior steps and initiate a new traversal accordingly.
% 具体来说，我们的方法提示模型逐行遍历表格的同时进行推理，每次遍历结束可以选择对之前的推理反思，并根据反思的结果开始下一次的遍历
% Specifically, \ours prompts LLMs to perform row-wise traversal reasoning, allowing for reflecting on prior inferences at the end of each traversal, and initiating the subsequent traversal based on the reflection.
% \ours缓解了Long CoT的两个限制
\ours alleviates two limitations of Long CoT:
% Low Cost：由于\ours根据表格逐行开展推理，且模型本身具备self-reflection能力，因此我们的方法无需训练，non-reasoning LLM仅根据提示就可以做到
(\emph{i})~Low Cost: Since \ours sequentially scales the reasoning length by row-wise traversals and the self-reflection capabilities are equipped in LLMs \cite{gu2025survey-judge,ai2025rethinking-reflection-pretraining}, \ours is training-free and can be implemented with non-reasoning large language models (non-RLLMs) through prompting. 
% High Reliability：\ours在推理时强制模型遍历所有行，令模型more attention to表格信息，可以减弱模型的幻觉
(\emph{ii})~High Reliability: By prompting the sequential traversal of all rows, \ours directs greater attention to tabular information thoroughly, thereby mitigating hallucination \cite{shi-etal-2024-trusting-Evidence-Hallucinate,chuang2024dola}.

% 实验
% 为了证明我们方法的有效性，我们在主流数据集WikiTQ, HiTab, TableBench上进行实验
To demonstrate the effectiveness of \ours, we conduct experiments on  WikiTableQuestions~\cite{pasupat-liang-2015-wikitq}, HiTab~\cite{cheng-etal-2022-hitab}, and TableBench~\cite{wu2024tablebench}. 
% 相比较同规模的RLLMs，我们方法使用non-reasoning LLMs平均提升%，证明了有效性
Compared to Long CoT on RLLMs, \ours achieves an average improvement of $4.3\%$ with non-RLLMs without training, validating its effectiveness. 
% 并且，我们的方法同样可以为推理模型带来进一步性能的提升，缓解了其在表格推理上的幻觉问题
Furthermore, \ours can also enhance the performance of RLLMs with an average improvement of $2.4\%$, mitigating their table content hallucination.
% 此外，我们在WikiTQ和TableBench上达到了同规模模型的SOTA，在HiTab上也同SOTA可比
Additionally, \ours achieves state-of-the-art (SOTA) results on WikiTableQuestions and TableBench with comparable models, and yields competitive results on HiTab. 
% 分析实验说明了\ours相较于长思维链，以更少的推理token达到了更强的性能，因为\ours缓解了推理模型过度思考、内容重复的问题
Analysis experiments reveal that \textbf{\ours with non-RLLMs outperforms Long CoT with fewer reasoning tokens, showing higher efficiency}.
% \cite{sui2025overthinking-survey,yin2025distillation-bottleneck,zeng2025revisiting-test-truly}.

% 我们的贡献如下
Our contributions are as follows:
\begin{enumerate}[nolistsep,leftmargin=*]
    % 1. 为了提升模型的表格推理能力，降低模型获得Long CoT的能力的成本，我们提出training-free的方法\ours，which提示模型逐行推理并反思
    % \item To reduce the cost of acquiring Long CoT abilities, we propose \ours, a training-free method that achieves sequential scaling and self-reflection by iterative row-wise traversals.
    % , scaling reasoning length.
    % 2. 为了缓解Long CoT在表格推理上的低可依赖性问题，\ours令模型在推理时逐行遍历完整的表格，以缓解对表格内容的幻觉
    % \item To alleviate the low reliability of Long CoT in table reasoning, \ours guides the model to traverse the table row-wise sequentially, thereby mitigating hallucinations of table content.
    % 我们提出了\ours，相较于Long CoT有着更低的cost，更高的可靠性
    \item We propose \ours, which achieves lower cost without training and higher reliability compared to Long CoT. 
    % 3. 我们在3个主流的表格推理数据集上进行实验，我们的方法平均超过推理模型的性能%,并且在WikiTableQuestions和TableBench上取得了同等规模下的SOTA，证明了有效性
    \item \ours on non-RLLMs outperforms Long CoT on RLLMs by an average of $4.3\%$ and achieves SOTA results among comparable models on WikiTableQuestions and TableBench, proving its effectiveness.
    % \ours with non-RLLMs相比较Long CoT用更少的推理token达到更好的性能，证明了\ours的高效性
    \item \ours with non-RLLMs outperforms Long CoT using fewer reasoning tokens, highlighting its higher efficiency.
\end{enumerate}

%% file: tex/3.methodology.tex
% 为了缓解Long CoT的High Cost and Low Reliability的限制，我们提出training-free的方法\ours
To mitigate the limitations of High Cost and Low Reliability in Long CoT, we propose \ours. 
% 如图所示，\ours通过令模型迭代逐行遍历表格增强表格推理能力
As illustrated in Figure~\ref{fig:method}, \ours enhances table reasoning capabilities by iterative row-wise traversals.
% 完整的prompt在附录中提供
The complete prompts are available in Appendix~\ref{subapp:Demonstrations of ours}.

\subsection{Overview}
% 我们输入提示I，问题Q和表格U，以及示例D，模型输出迭代逐行遍历表格直到得到答案的过程
Given an instruction $I$, a question $Q$, a table $U$ composed of $M$ rows and $N$ columns, and in-context demonstrations $D$, the model outputs a step-by-step reasoning process that iteratively traverses the table in the sequential row order until the final answer is derived.
% 形式化地，R;A = M(Q, U, I, {Q^i, U^i, R^i, A^i})，其中，M为语言模型，D = {Q^i, U^i, R^i, A^i}，R;A代表推理过程和答案的连接
% R为推理过程，A为最终答案
Formally, $R;A = \mathcal{F}(I, Q, U, D)$, where $\mathcal{F}$ is the LLM, and $R;A$ denotes the concatenation of the reasoning process $R$ and the answer $A$.
% $R$ represents the reasoning process, and $A$ is the final answer. 
% 我们采用markdown的格式表示表格，follow前人工作
We represent the table in Markdown format, following previous works \cite{wang2024chainoftable,zhang2024flextaf,yu2025table-critic}. 
% 下面我们分别介绍在\ours的推理过程$R$中的两个重要因素
We now introduce the two key factors in the reasoning process $R$ in \ours.

\subsection{Traversal}
% 我们创新性提出逐行遍历的推理方法，which不同于前人工作中勇于表格推理的CoT方需要模型自主决定每一步推理对应的子表
% We innovatively propose a row-wise traversal reasoning pattern, which is different from previous CoT methods for table reasoning \cite{wang2024chainoftable,ye2023dater,ji2024tree-of-table}.
% We introduce a novel traversal reasoning method adopting the row as the traversal unit, unlike previous CoT methods \cite{wang2024chainoftable,ye2023dater,ji2024tree-of-table} for table reasoning that necessitate the autonomous selection of sub-tables at each step.
% 我们首先介绍推理中逐行遍历的过程
We first detail the traversal reasoning adopting the row as the traversal unit in \ours.
% 具体来说，模型根据问题、之前的推理和当前行，对当前行中的信息判别和问题的相关性并推理得到中间结果
Specifically, the model assesses the relevance of information within the current row and infers intermediate results according to the question and prior inference.
% 形式化地，R_i; A_i = R_{i,1};A_{i,1}, R_{i,2};A_{i,2}, ..., R_{i,M};A_{i,M}
Formally, $R_i; A_i = R_{i,1};A_{i,1}, R_{i,2};A_{i,2}, ..., R_{i,M};A_{i,M}$.
% 其中，R_i代表第i次遍历的推理过程，A_i为第i次遍历得到的结果
$R_i$ represents the reasoning process of the $i$-th traversal, and $A_i$ is the result obtained in the $i$-th traversal.
% R_{i,j}代表第i次遍历对第j行表格的推理，A_{i,j}为对应的中间结果
$R_{i,j}$ denotes the reasoning over the $j$-th row of the table during the $i$-th traversal, and $A_{i,j}$ is the corresponding intermediate result. 
% \ours利用表格天然的结构特征，将问题求解分解为细粒度的分步推理，每一步对应一行
\ours leverages the inherent structural features of tables by decomposing the problem-solving into fine-grained, step-by-step reasoning, with each step corresponding to a row. 
% 通过累积每一行的中间结果A_{i,j}，得到一次遍历后的结果A_i
By accumulating the intermediate results $A_{i,j}$ from each row, we obtain the result $A_i$ after one traversal.
% 通过逐行的推理，不仅scale了推理长度，而且强制模型关注全部的表格内容缓解了幻觉
The row-wise traversal not only brings the reasoning length scaling but also mitigates hallucination by forcing the model to attend to the entire table content.
% 我们还在4.3.5节讨论了采用了其他遍历单位和\ours的对比
We also discuss comparisons with adopting other traversal units and \ours in \S\ref{subsec:traversal unit}.

\subsection{Iteration}
% 在本小节，我们具体介绍\ours中模型如何迭代地遍历表格
% This subsection details how our model iteratively traverses tables.
% 迭代地遍历具体表现为每次遍历结束，模型可以选择继续推理，因为涉及多行依赖或多跳的问题无法在一次遍历中得到答案
The iteration process allows the model to continue reasoning after a traversal, which is necessary for multi-hop questions that cannot be answered in a single traversal.
% 模型也可以选择在遍历结束后对之前的推理进行反思，并根据反思的内容重新遍历表格直到得到最终答案
Also, the model can choose to reflect on the previous reasoning after a traversal and subsequently revisit the table based on the reflection until the final answer is obtained.
% 形式化地，R;A = R_1;A_1, R_2;A_2, ...R_T;A_T
Formally, the iterative reasoning process can be represented as $R;A = R_1;A_1, R_2;A_2, ..., R_T;A_T$, where $T$ is the total number of traversals.
% 其中，T为总遍历次数
% 我们没有固定模型遍历表格的次数，而是由模型在推理时自主决定，当认为得到最终答案时停止
% Instead of predefining $T$ in the prompt, the model autonomously decides when to stop during inference upon reaching a final answer.
Rather than predefining $T$ in the prompt, the model dynamically decides to terminate inference when the final answer has been obtained.
% 我们在4.3.2节详细分析了模型多次遍历表格的情况及分布
We provide a detailed analysis of the iterative table traversals in \S\ref{subsub:table traversal counts}.
We also provide case study for iterative traversals in Appendix~\ref{subapp:Case study of ours with iterative traversals}.

%% file: tex/4.experiment.tex
\subsection{Experimental Setup}
\paragraph{Dataset}
% 我们follow前人工作，使用WikiTableQuestions, HiTab和TabelBench的popular表格推理数据集上评测我们方法的性能
\ours is evaluated on three widely used table reasoning datasets: WikiTableQuestions~\cite{pasupat-liang-2015-wikitq}, HiTab~\cite{cheng-etal-2022-hitab}, and TableBench~\cite{wu2024tablebench}, following previous works \cite{jiang2024sscot,cao2025tablemaster,li-etal-2025-graphotter}. 
% WikiTQ是表格推理的主流表格QA数据集，需要模型根据表格回答问题
WikiTableQuestions is a mainstream table-based question answering dataset.
% that requires models to reason over tabular content.
% HiTab关注于hierarchical表格，对模型理解表格展现出挑战
HiTab focuses on hierarchical tables, challenging models to comprehend complex structural relationships.
% TableBench是综合的、具有挑战的数据集，包含多样的问题类型和topic
TableBench presents a challenging benchmark covering diverse question types and topics.

\paragraph{Models}
% 对于语言模型，我们采用Llama3.1-8B-Instruct (Llama3.1-8B), Llama3.3-70B-Instruct (Llama3.3-70B)，Qwen2.5-7B-Instruct (Qwen2.5-7B)和Qwen2.5-32B-Instruct (Qwen2.5-32B)
(\emph{i})~For \textbf{non-RLLMs}, we utilize Llama3.1-8B-Instruct (Llama3.1-8B), Llama3.3-70B-Instruct (Llama3.3-70B) \cite{dubey-etal-2024-llama3.1}, Qwen2.5-7B-Instruct (Qwen2.5-7B), and Qwen2.5-32B-Instruct (Qwen2.5-32B) \cite{yang2024qwen25}. 
% 对于推理模型，我们采用对应规模的DeepSeek-R1-Distill-Llama-8B (R1-Llama-8B), DeepSeek-R1-Distill-Llama-70B (R1-Llama-70B), DeepSeek-R1-Distill-Qwen-7B (R1-Qwen-7B)和DeepSeek-R1-Distill-Qwen-32B (R1-Qwen-32B)
(\emph{ii})~For \textbf{RLLMs}, we employ the corresponding-sized DeepSeek-R1-Distill-Llama-8B (R1-Llama-8B), DeepSeek-R1-Distill-Llama-70B (R1-Llama-70B), DeepSeek-R1-Distill-Qwen-7B (R1-Qwen-7B), and DeepSeek-R1-Distill-Qwen-32B (R1-Qwen-32B) \cite{guo2025deepseek-r1}. 
% 其中，我们没有采用DeepSeek-R1-Distill-Qwen-7B的base model Qwen2.5-Math-7B，是因为这一模型主要适用于数学任务，在表格推理任务上的性能较差
We exclude Qwen2.5-Math-7B, which is the base model of R1-Qwen-7B, due to its primary focus on solving mathematical tasks, resulting in suboptimal performance on the table reasoning task \cite{yang2024qwen25-math}.

\paragraph{Metric}
% 对于WikiTableQuestions和HiTab，我们采用准确率作为评价指标，follow前人工作
For WikiTableQuestions and HiTab, we adopt accuracy as the evaluation metric, following prior works \cite{pasupat-liang-2015-wikitq,cheng-etal-2022-hitab}. 
% 准确率衡量了模型生成和gold answer完全匹配的答案的能力
Accuracy measures the ability of models to generate answers that exactly match the gold answers. 
% 对于TableBench，我们采用Rouge-L，follow前人工作
For TableBench, we use Rouge-L~\cite{lin-2004-rouge}, consistent with the previous research \cite{wu2024tablebench}. 
% Rouge-L通过最长公共子序列衡量模型生成的答案的质量，兼顾考虑了Precision和Recall
Rouge-L evaluates the quality of generated answers based on the longest common subsequence, considering both precision and recall.

\begin{table*}[t]
    \centering
    \small

\input{tab/main}

    \caption{ 
        % 我们的方法和基线的性能对比
        Performance comparison between \ours and baselines, where WikiTQ and HiTab use accuracy as the evaluation metric and TableBench uses Rouge-L.
        % WikiTQ指WikiTableQuestions
        WikiTQ refers to WikiTableQuestions.
        % 每个数据集上同一个规模的模型上最好的结果被标志加粗
        For each dataset, the highest performing result among models of the same scale is \textbf{bolded}.
        % 括号中的数字为\ours相比基线的性能提升
        Performance gain compared to baselines is highlighted with \green{(green)}.
    }
    \label{tab:main}
\end{table*}

\paragraph{Baselines}
% 对于我们的方法，我们采用one-shot的prompt，为了令模型详细了解如何逐行遍历表格
\ours employs the one-shot and zero-shot prompts to enable \textbf{non-RLLMs} and \textbf{RLLMs} to perform iterative row-wise traversals, respectively (prompts in Appendix~\ref{subapp:Demonstrations of ours}). 
We do not use demonstrations for RLLMs due to the performance degradation using the few-shot prompt observed in Appendix~\ref{subapp:Long CoT with few-shot prompt} \cite{guo2025deepseek-r1,zheng2025curse-cot-limitations-icl}.
% 因此，我们将我们的方法和以下方法进行比较
We compare \ours with the following methods:
\begin{itemize}[nolistsep,leftmargin=*]
    % CoT with language model （one-shot）：我们采用了和我们方法相同的示例，提示语言模型think step by step再输出答案
    \item Short CoT: We prompt \textbf{non-RLLMs} to engage in step-by-step reasoning with the one-shot prompt, which uses the same demonstration as \ours.
    % CoT with reasoning model （zero-shot）：我们提示推理模型回答问题。由于我们采用的推理模型不具备in-context learning的能力，加上示例反而会带来性能的下降（见附录），所以我们采用zero-shot的prompt
    \item Long CoT: We utilize the zero-shot prompt for \textbf{RLLMs}.
    % 其他表格推理的前人工作：我们在附录将我们的方法和先进的前人工作比较
    \item Previous table reasoning works: We compare \ours with existing table reasoning methods with comparable models.
\end{itemize}

\begin{table}[t]
    \centering
    \small
    \input{tab/SOTA_comparison}
    \caption{ 
        % 我们的方法和基线的性能对比
        Performance comparison between \ours and SOTA methods with similar scale models.
    }
    \label{tab:SOTA_comparison}
\end{table}

\subsection{Main Results}
% 表1展现了我们的方法使用不同模型在不同数据集上和基线的对比
Table~\ref{tab:main} presents a comparison between \ours and baselines using different models across datasets.
% 可以发现，\ours一致且显著地超越了推理模型的性能，平均提升%
\ours, using non-RLLMs consistently and significantly outperforms Long CoT with RLLMs, achieving an average improvement of $4.3\%$, demonstrating its effectiveness.
% 并且，\ours能为推理模型带来进一步的提升，平均提升3%，说明\ours很好地缓解了Long CoT的限制
Furthermore, \ours yields an average increase of $2.4\%$ in the performance of RLLMs, indicating its effectiveness in mitigating the limitations of Long CoT.
% 我们还发现
We also observe that:

\paragraph{\ours outperforms baselines consistently.}
% 我们的方法超越Long CoT，因为我们强制模型按照行序遍历表格，避免了Long CoT因为输出过长，对表格信息的lose从而产生幻觉
\ours surpasses Long CoT primarily because it enforces the row-wise traversals, alleviating hallucinations in Long CoT \cite{zhang2023sirens-Hallucination-survey,shi-etal-2024-trusting-Evidence-Hallucinate,Liu2025hallucination-long}.
% 我们在附录结合示例详细分析了为什么\ours with 推理模型可以超越Long CoT
% In Appendix~\ref{subapp:ours with RLLMs outperforms Long CoT}, we provide a detailed analysis with examples illustrating why \ours with RLLMs can outperform Long CoT.
% 我们的方法能够超越Short CoT，是因为它逐行遍历表格进行细粒度推理，降低了每一步的推理难度，而且不容易遗漏相关信息
Compared to Short CoT, \ours achieves superior performance through fine-grained, row-wise reasoning, thereby reducing the complexity of individual reasoning steps and minimizing the risk of overlooking relevant details \cite{snell2024scaling-test-time,wang2024chainoftable}.

% 并且，我们将我们的方法和前人方法在WikiTQ和HiTab上进行比较，如表2和表3所示
We also compare \ours with SOTA methods on three datasets, as shown in Table~\ref{tab:SOTA_comparison}.
% 受限于空间，我们在附录详细介绍SOTA方法，并将我们的方法和更多前人工作比较
Due to space constraints, detailed comparisons with prior works are provided in Appendix~\ref{subapp:Comparison with previous methods}.
% 可以发现，我们的方法超过了使用同等规模甚至更大模型的前人工作的性能，进一步证明了我们方法的有效性
\ours gets SOTA results on WikiTQ and TableBench and is comparable with the SOTA method on HiTab, highlighting its effectiveness.
% 我们的方法没有像前人方法那样专门根据嵌套表格进行改进，所以没有在HiTab上也达到SOTA
The comparable performance on HiTab can be attributed to the fact that \ours does not incorporate specific enhancements for hierarchical tables, unlike previous methods \cite{zhao-etal-2023-parser,jiang2024sscot,li-etal-2025-graphotter}.

\begin{table*}[t]
    \centering
    \small
    \input{tab/ablation}
    \caption{ 
        The ablation results of \ours campared with reasoning with one single table traversal (denoted as \textit{w/o Iteration}) and reasoning without table traversal (denoted as \textit{w/o Traversal}).
        % 每个数据集上同一个规模的模型上最好的结果被标志加粗
        For each dataset, the highest performing result with the same model is \textbf{bolded}.
    }
    \label{tab:ablation}
\end{table*}

\paragraph{\ours improves performance across varying models.}
% 我们的方法在不同规模的模型上均得到显著提升，且均超过对应推理模型的性能
% \ours consistently demonstrates substantial performance improvements across varying models and scales without training.
% 我们的方法可以显著地提升不同模型的表格推理能力 without training，无论是推理模型还是non-reasoning LLMs
\ours significantly enhances the table reasoning capabilities of various non-RLLMs and RLLMs without training. 
% 而对表格内容的幻觉问题在不同规模的推理模型上普遍存在，所以我们的方法可以在推理模型上得到一致的提升
% Furthermore, given the prevalent issue of table content hallucination across RLLMs, \ours yields consistent improvements on these models.
% \ours with RLLMs有时未能超过\ours with non-reasoning LLMs，因为虽然我们缓解了RLLMs的幻觉问题，但RLLMs有着过度思考，内容重复等问题，而这些问题在non-reasoning LLMs上很微弱
\ours with RLLMs does not outperform \ours with non-RLLMs consistently because, while we mitigate hallucination in Long CoT, they exhibit problems such as overthinking, which are less pronounced in non-RLLMs \cite{yin2025distillation-bottleneck,zeng2025revisiting-test-truly}.
% 同时，R1-Qwen-7B在HiTab上的性能未超过Qwen2.5-7B，因为R1-Qwen-7B的Base model是Qwen2.5-Math-7B，which主要用于解决数学问题，不同于其他模型的base model是general models
Additionally, R1-Qwen-7B does not outperform Qwen2.5-7B on HiTab, as its base model, Qwen2.5-Math-7B, is optimized for mathematical reasoning, unlike the general base models of others \cite{yang2024qwen25-math}.

% \paragraph{Datasets}
% % 我们的方法在HiTab上的提升相较于其他数据集上的提升更显著
% \ours yields more pronounced improvements on HiTab compared to other datasets. 
% % 因为模型对于嵌套表格的理解比较困难，而我们的方法通过逐行遍历更加细节地理解了表格
% This can be attributed to the greater complexity of hierarchical tables for table understanding, where our row-wise traversal facilitates a more fine-grained comprehension of table structure and content. 

\subsection{Ablation Experiments}
% 为了证明\ours的有效性，我们在三个数据集上构建了消融实验，如表所示
To demonstrate the effectiveness of \ours, we conduct ablation experiments on three datasets, as shown in Table~\ref{tab:ablation}. 
% 消融实验使用的prompt在附录中提供
The prompts used in the ablation experiments are provided in Appendix~\ref{subapp:Prompts for ablation experiments}.

\paragraph{Effectiveness of Iteration}
% 为了证明\ours中迭代遍历表格的有效性，我们提示模型只能遍历一次表格来进行实验
To validate the effectiveness of iterative reasoning in \ours, we prompt the model to perform only a single table traversal. 
% 可以发现，移除迭代后的性能相比较\ours一致地下降，证明了迭代遍历有效帮助模型进行深度推理以及反思
The results indicate a consistent performance decrease compared to \ours when iteration is removed, demonstrating that iterative traversal effectively aids the model in exploration and reflection. 
% 而只用一次遍历无法很好地解决所有表格问题
Also, a single traversal is insufficient to adequately address all table reasoning questions.

\paragraph{Effectiveness of Traversal}
% 为了证明\ours中遍历表格的重要性，我们提示模型迭代地反思，而不是迭代的逐行遍历表格
To demonstrate the importance of traversal in \ours, we prompt LLMs to iteratively reflect instead of iteratively traversing the table. 
% 可以发现，模型不再遍历表格后，性能下降显著，说明了遍历表格通过细粒度推理，缓解幻觉有效增强了表格推理能力
% The significant performance drop observed when the model no longer traverses the table highlights the benefit of table traversal for table reasoning capabilities.
The significant performance decline observed underscores that traversing the table, through scaling reasoning length and mitigating hallucinations of tabular content, effectively enhances table reasoning.

\subsection{Analysis Experiments}
% 我们主要选取Llama3.1-8B进行后续分析实验，受限于篇幅
We primarily select Llama3.1-8B and R1-Llama-8B for subsequent analysis experiments due to their high reasoning efficiency and space limitations. 

\begin{figure}[t]
\centering
\input{fig/error}
% \vspace{-0.5em}
\caption{
    % Long CoT因为呈现出的错误类型而不如\ours
    Long CoT underperforms \ours due to the error types, with their distribution.
}
\label{fig:error}
\end{figure}
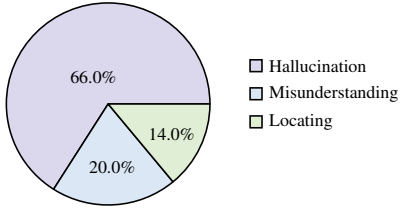

\subsubsection{Why \ours Outperforms Long CoT?}
\label{subsec:ours Outperforms Long CoT}
% 为了更具体地分析为什么我们方法的性能超过Long CoT，我们随机选取了我们的方法使用Llama3.1-8B在WikiTQ上的做对，但Long CoT使用R1-Llama-8B没做对的instance，手动分析了原因，比例如图所示
To explore the superior performance of \ours over Long CoT, we conduct an error analysis on WikiTQ instances where \ours with Llama3.1-8B succeeds while Long CoT with R1-Llama-8B fails. 
We also explore why \ours with RLLMs outperforms Long CoT in Appendix~\ref{subapp:ours with RLLMs outperforms Long CoT}.
Figure~\ref{fig:error} illustrates the identified error categories on sampled $50$ instances, which are detailed below.
We provide the cases of each error category in Appendix~\ref{subapp:Case study of ours compared with Long CoT}.

\begin{figure}[t]
\centering
\input{fig/multi_traversal}
\caption{
    % 我们的方法遍历多次表格的原因，及其比例
    The distribution of reasons for iterative traversals in \ours on sampled $60$ instances from three datasets.
}
\label{fig:multi_traversal}
\end{figure}
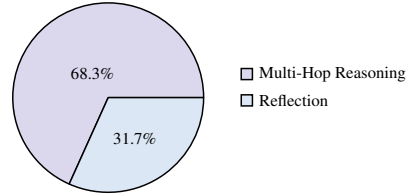

% Hallucination指的是模型记错了表格中的信息，出现在推理中的表格内容和输入的不一致，比如混淆了不同行的内容
(\emph{i})~\textbf{Hallucination} refers to the model incorrectly recalling tabular information, leading to inconsistencies between the table input and the generated reasoning, such as cross-row confusion and relevant information omission. 
% Long CoT幻觉严重，是因为随着输出的推理链条的变长，模型对表格内容的丢失导致了
% Long CoT exhibits a higher frequency of hallucinations, due to its frequent shifts in reasoning paths, which cause the model to forget input details \cite{Liu2025hallucination-long}. 
Long CoT suffers from severe hallucinations, primarily due to the increasing loss of tabular content as the reasoning chain lengthens \cite{Liu2025hallucination-long}. 
% 而我们的方法按照行序顺序遍历，符合注意力机制，减轻了这种现象
Conversely, \ours performs row-wise traversals sequentially, guides greater attention to the table content, which mitigates this issue \cite{yin-etal-2020-tabert,badaro-etal-2023-transformers-table-survey}.
% Overlooking指的是模型没有关注到表格中的问题相关信息
% (\emph{ii})~\textbf{Overlooking} refers to failing to focus on the relevant information in the table. 
% % 由于Long CoT的推理顺序较为自由，在不同位置的cell间跳转很容易没有覆盖到所有表格，进而忽视了相关内容
% The flexible reasoning order and frequent transitions between cells in Long CoT increase the susceptibility to overlooking crucial data.
% % 而我们的方法要求遍历表格所有内容，减轻了这种现象
% In contrast, \ours requires a complete traversal of the table, thus alleviating this issue.
% Misunderstanding指的是模型由于误解问题而产生错误的推理，这是蒸馏模型带来的问题
(\emph{ii})~\textbf{Misunderstanding} denotes the misinterpretation of the question, which is a common challenge for distilled models \cite{banerjee2024llms-hallucinate-live,yin2025distillation-bottleneck}.
% Linking指的是模型根据问题定位到了表格中错误的位置
(\emph{iii})~\textbf{Locating} refers to incorrectly identifying the relevant table location for the given question. 
% 说明我们的方法相比Long CoT有着更高的依赖
Therefore, \ours demonstrates a higher reliability compared to Long CoT.

\begin{figure*}[ht]
\centering
\includegraphics[width=.95\linewidth]{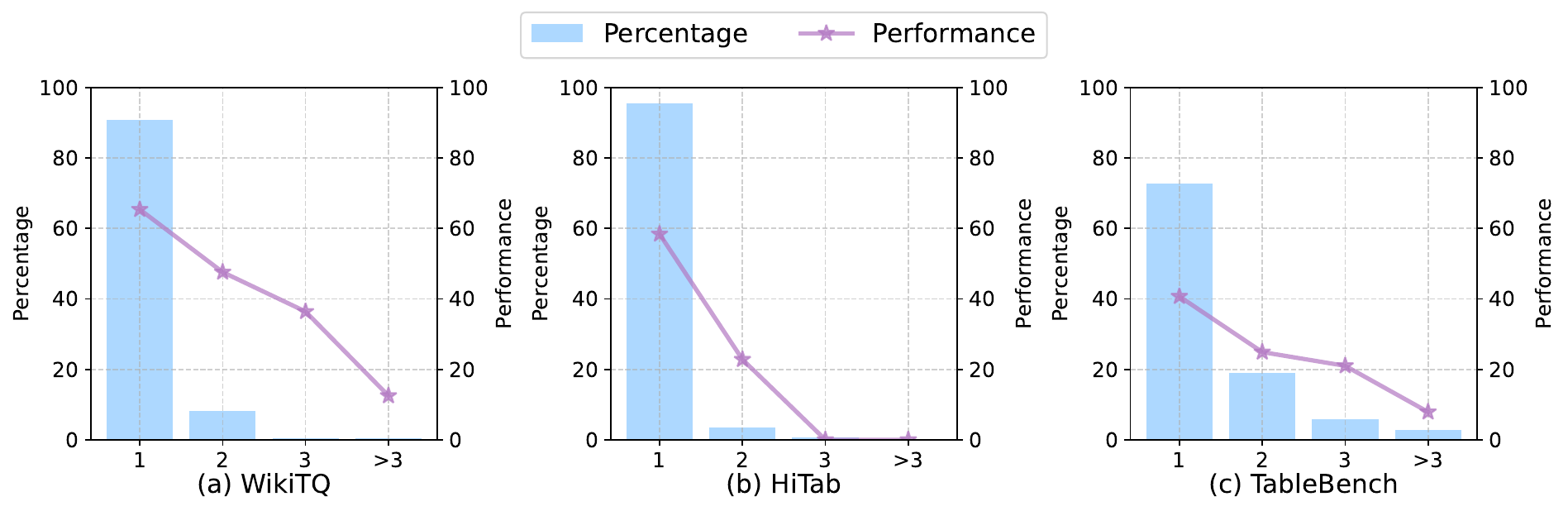}
\vspace{-0.5em}
\caption{
    % 在3个数据集上，我们方法遍历表格次数的分布和相应的性能
    The distribution of table traversal counts and the corresponding performance of \ours on three datasets with Llama3.1-8B.
}
\label{fig:traversal_num}
\end{figure*}

% \begin{figure*}[htbp]
% \centering
% \input{fig/traversal_num}
% \caption{
%     % 在3个数据集上，我们方法遍历表格次数的分布和相应的性能
%     The distribution of table traversal counts and the corresponding performance of \ours on three datasets.
% }
% \label{fig:traversal_num}
% \end{figure*}

\subsubsection{How does the number of traversals affect \ours?}
\label{subsub:table traversal counts}
% 为了分析我们的方法多次遍历表格的原因季分布，我们分别从3个数据集随机抽取了20条我们方法大于一次地遍历表格的数据并分析其多次遍历的原因，如图所示
To examine when \ours requires iterative traversals, we randomly select $20$ instances from each dataset on Llama3.1-8B where \ours traverses the table more than once and investigate the reasons, as shown in Figure~\ref{fig:multi_traversal}. 
% 我们首先详细介绍这3个多次遍历的原因，对应的例子见附录
We provide a detailed explanation of the reasons below, with examples provided in Appendix~\ref{subapp:Case study of ours with iterative traversals}.
% 1. Multi-Step Reasoning：问题本身需要多次遍历表格才能解决，比如跨行依赖等
(\emph{i})~\textbf{Multi-Hop Reasoning}: The inherent complexity of certain questions demands iterative table traversals to derive the solution, particularly when addressing \textit{cross-row dependencies}.
% 2. Reflection：模型在遍历结束时对之前的推理进行反思，并据此开启新的遍历推理以检查或纠错
(\emph{ii})~\textbf{Reflection}: The model reflects on its prior reasoning upon completing a traversal and initiates new reasoning passes accordingly.
% 这表明，\ours很好地激发了模型深度推理和并行推理的能力
This demonstrates that \ours with non-RLLMs equips the capacities of extending reasoning and self-reflection on table reasoning.
% 这表明非推理模型也存在self-reflection的能力，而我们的方法综合地激发了模型的表格推理能力，包括self-reflection的能力
% This observation suggests an inherent capacity of self-reflection even in non-reasoning LLMs \cite{ai2025rethinking-reflection-pretraining}, and \ours effectively elicits comprehensive table reasoning capabilities, including this self-reflective ability.
% 3. Summary：在进行必要的推理后，重新遍历表格总结答案。
% (\emph{iii})~\textbf{Summarization}: Following the necessary reasoning steps, the model re-traverses the table to generate a concise summary of the answer.

% 并且为了分析遍历次数对我们方法的影响，我们统计了在使用Llama3.1-8B时遍历表格次数的分布以及对应的性能，如图所示
Additionally, to assess the impact of traversal count on performance, we report the distribution of table traversal counts and the corresponding performance when using Llama3.1-8B, as depicted in Figure~\ref{fig:traversal_num}.
% 观察图3，我们发现
We observe that:
% 1. 即使我们没有显式提示模型遍历几次表格，但模型会按需遍历多次。在更具有挑战性的TableBench上我们的方法倾向于遍历次数更多
(\emph{i})~On the more challenging TableBench dataset, \ours tends to perform more traversals as required.
% 2. 随着遍历次数的增多，我们方法的性能下降。因为一方面需要多次遍历的问题本身更难，一方面多次遍历表格容易造成在得到最终答案之前超出token限制
(\emph{ii})~Increasing traversal counts correlate with a decrease in the performance of \ours, due to the inherent difficulty of questions necessitating iterative traversals and the potential for exceeding token limits during such processes.

% 推理token的统计
\subsubsection{How does reasoning length affect table reasoning capabilities?}
% 为了探索推理长度对表格推理性能的影响，我们分别统计了不同方法的正确推理平均长度和错误的推理平均长度，如图所示
To investigate the impact of reasoning length on table reasoning performance, we calculate the average number of tokens used in correct and incorrect reasoning on WikiTQ, as shown in Figure~\ref{fig:token_num}. 
% 可以发现
The results reveal that:\begin{figure}[t]
    \centering
    \includegraphics[width=.9\linewidth]{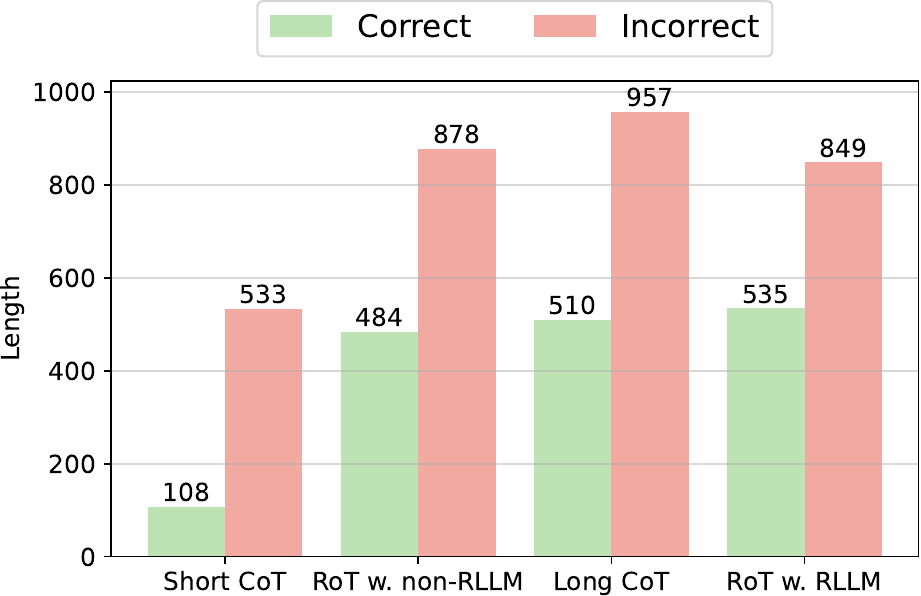}
    % \vspace{-1.5em}
    \caption{
    % 在3个数据集上，我们方法和基线正确推理平均长度和错误的推理平均长度的对比
    Comparison of average reasoning lengths for correct and incorrect inferences across three datasets on WikiTQ with Llama3.1-8B (denoted as w. non-RLLM) and R1-Llama-8B (denoted as w. RLLM).
    }
    \label{fig:token_num}
\end{figure}

% 1. \ours with non-RLLM使用较少的token实现了表格推理能力的提升，说明了\ours的高效性
(\emph{i})~\ours with non-RLLM achieves improved table reasoning with fewer tokens compared to Long CoT, demonstrating its efficiency. 
% 因为\ours允许模型自己决定迭代次数，且non-RLLM没有专门在Long CoT数据上训练过，所以缓解了Long CoT中过度思考、重复输出相同内容的情况
 \ours allows the model to dynamically determine the number of iterations and non-RLLMs are not specifically trained on Long CoT data, therefore, \ours mitigates overthinking prevalent in Long CoT \cite{yin2025distillation-bottleneck}. 
% 并且，同样使用RLLM，\ours相比Long CoT在错误的推理上更短，也是因为\ours通过更加关注表格减弱了模型对表格内容的幻觉，因此减少了无效反思的频率
Additionally, when using the same RLLM, \ours exhibits shorter incorrect reasoning compared to Long CoT, since \ours, by focusing more intently on the table, reduces model hallucinations regarding table content, thereby decreasing the frequency of ineffective reflections, as discussed in Appendix~\ref{subapp:ours with RLLMs outperforms Long CoT} \cite{shi-etal-2024-trusting-Evidence-Hallucinate,qin2025backtrackback-when}.
% 2. 使用相同的模型，\ours相比对应的CoT，正确的推理更长
(\emph{ii})~Using the same model, \ours produces longer correct reasoning compared to its corresponding CoT baseline.
% 因为逐行遍历表格能够更细粒度地推理，增加了推理长度的同时带了性能提升
This is because the row-wise table traversal enables more fine-grained reasoning, leading to increased reasoning length and improved performance \cite{qian2025fino1}.
% 因为整体上我们方法的推理长度长于Short CoT，短于Long CoT，且我们的方法性能最好
% Specifically, the average reasoning length of \ours lies between that of Short CoT and Long CoT, yet it achieves the highest overall performance.
% % 一方面，因为逐行遍历表格能够更细粒度地推理，而且减少了遗漏相关信息的风险
% This can be attributed to two factors. 
% First, row-wise table traversals facilitate more granular reasoning and mitigate the risk of overlooking pertinent information.
% % 另一方面，我们使用的模型没在长推理数据上专门训练过，可以按照问题需求灵活变化推理长度
% Second, non-reasoning LLMs are not explicitly trained on Long CoT data, enabling them to adapt the reasoning length flexibly based on question complexity.
% % 2. inference-time scale可以激发模型的表格推理性能，但过多的推理token也会带来性能下降。
% (\emph{ii})~Inference-time scaling can enhance the table reasoning capability, however, excessive reasoning tokens lead to performance degradation. 
% % Long CoT错误的推理长度格外的长，说明了频繁切换推理路径容易导致遗漏表格中的相关信息，并且过度思考、内容重复、错误累积等现象严重
% The notably longer reasoning length observed in incorrect predictions of Long CoT suggests that frequent transitions between reasoning paths can result in the hallucination and omission of relevant table information \cite{he2025detect-errors,arcuschin2025CoT-not-Faithful,yin2025distillation-bottleneck}.
\setcounter{figure}{7}
\begin{figure*}[ht]
    \centering
    \includegraphics[width=1\linewidth]{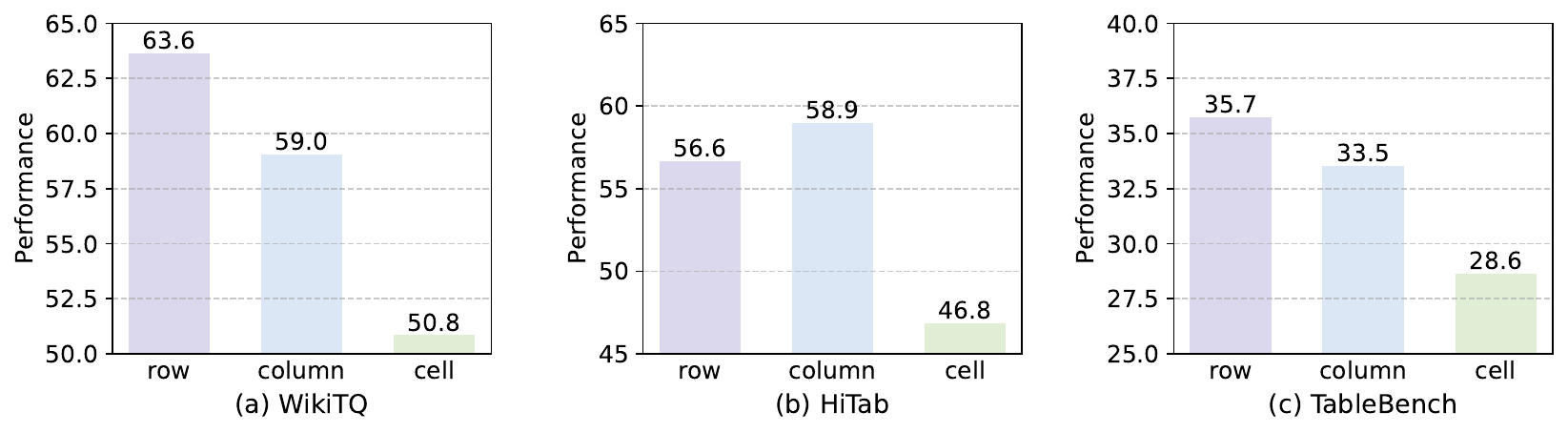}
    \vspace{-1.5em}
    \caption{
    % 在3个数据集上，我们方法使用不同遍历单位的性能比较
    Comparison of \ours traversing the table with different units across three datasets.
    }
    \label{fig:unit_performance}
\end{figure*}

\addtocounter{figure}{-2}
\begin{figure}[t]
    \centering
    \includegraphics[width=.9\linewidth]{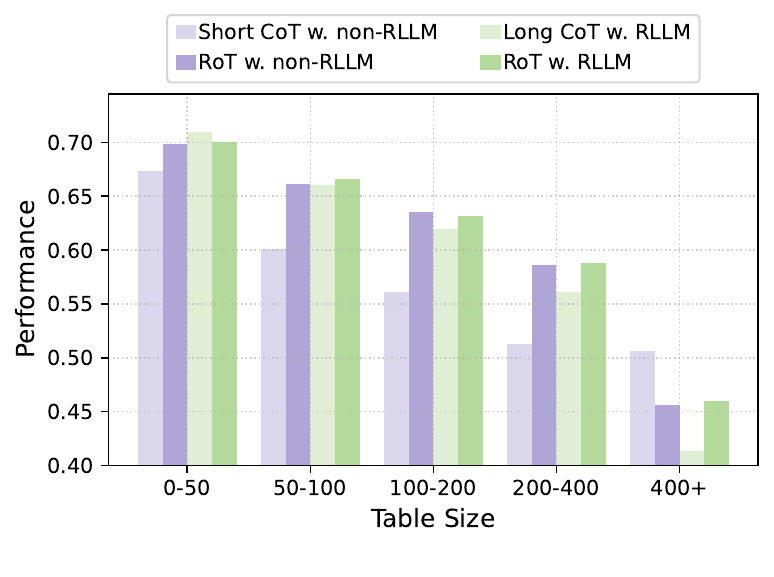}
    % \vspace{-1.5em}
    \caption{
    % 在3个数据集上，我们方法和基线在不同表格尺寸上平均性能的比较
    The comparison of the average performance of \ours and baselines on different table sizes in WikiTQ with Llama3.1-8B and R1-Llama-8B.
    }
    \label{fig:table_size}
\end{figure}
\setcounter{figure}{8}

% 我们的方法随着表格尺寸的变化如何变化
\subsubsection{How does \ours change with table size?}
% 为了分析我们的方法在不同表格尺寸上的性能变化，我们比较了我们的方法和基线使用Llama3.1-8B在不同表格尺寸上的性能，其中表格尺寸是表格行数和列数的乘积，如图所示
To evaluate the performance of \ours relative to baselines across varying table sizes, we analyze the performance of Llama3.1-8B and R1-Llama-8B on tables of different sizes in WikiTQ, defined as the product of the number of rows and columns (Figure~\ref{fig:table_size}).
% 可以发现
The key observations are as follows:
% 1. 我们的方法在不同尺寸上的性能整体上优于基线
(\emph{i})~Overall, \ours outperforms the baselines across table sizes.
% 2. 我们的方法随着表格尺寸变大，性能变化较为稳定，但整体呈现下降趋势
(\emph{ii})~While exhibiting a general downward trend, the performance of \ours demonstrates relative stability with increasing table size.
% 我们的方法由于需要逐行遍历表格进行细粒度的推理，表格行数过多容易在还未得到答案时超出token限制
The row-wise traversals could lead to exceeding the token limit when the number of rows becomes excessively large before a response is generated.
% 而推理模型由于表格尺寸较大，推理步骤和推理路径增多，也容易超出token限制
Long CoT suffers from an increased number of reasoning steps with larger tables, elevating the risk of hallucinating relevant information and surpassing token limits more significantly \cite{zeng2025revisiting-test-truly,sui2025overthinking-survey}.
% 而普通CoT的方法虽然不容易超出token限制，但推理的粒度过大，容易遗漏或找错相关的表格内容
Short CoT, while less susceptible to token limit issues, could overlook relevant table information due to its coarser reasoning granularity and miss self-reflection reasoning \cite{snell2024scaling-test-time,zhang2025why-cotprompt-not-work}.

% 不同单位的遍历方法
\subsubsection{How does the traversal unit affect \ours?}
\label{subsec:traversal unit}
% 为了探索遍历单位对我们方法的影响，我们分别采用行、列和cell使用Llama3.1-8B进行实验，结果如图所示
To investigate the effect of traversal units on \ours, we conduct experiments using rows, columns, and individual cells as traversal units across three datasets with Llama3.1-8B.
% 其中，逐行遍历是我们主实验采用的单位
Row-wise traversal is adopted as the default setting in the main experiments.
% 结果显示
The results indicate the following:

% 1. 在WikiTQ和TableBench上我们的方法都是以行为单位的性能最高
(\emph{i})~On WikiTQ and TableBench, row-wise traversal achieves the best performance. 
% 因为相比按列遍历，逐行遍历更符合模型的注意力机制，便于模型将注意力集中在同一行的各个单元格上
Compared to column-wise traversal, row-wise traversal better aligns with the attention mechanism, enabling more effective focus on all cells within the same row \cite{yin-etal-2020-tabert,liu-etal-2024-rethinking-tabular}. 
% 而按cell遍历性能下降显著，是因为按cell遍历的推理粒度过于小，且大量的cell是问题无关的，冗余的推理步骤增加了误差累积的风险
Cell-wise traversal resulted in a significant performance decrease, due to its overly fine-grained reasoning granularity and the presence of numerous irrelevant cells, which introduce redundant reasoning steps and increase the risk of error accumulation \cite{jin-etal-2024-impact-Reasoning-Step,chen2024Boundary,patnaik2024cabinet}.
% 2. 相反，在HiTab上我们的方法按列遍历的性能最好
(\emph{ii})~Column-wise traversal yields superior performance on HiTab. 
% 考虑到HiTab中的cell可能对应多个嵌套的header，逐行遍历时需要对应每个cell的内容到多个header，而逐列遍历时header的信息已经包含到这一列中了，所以按列遍历的性能更好
% HiTab中所有的表格都具有hierarchical row headers，93.1%的表格具有hierarchical column headers，相对较少
In HiTab, all tables include hierarchical row headers, while hierarchical column headers are present in $93.1\%$ of the tables, a relatively less frequent occurrence \cite{cheng-etal-2022-hitab}. 
% 因此，其中每行都每个cell都对应hierarchical row headers，逐行遍历时需要对应每个cell的内容到多个row headers，而逐列遍历时header的信息已经包含到这一列中了，所以按列遍历的性能更好
Consequently, each cell in a row corresponds to hierarchical row headers. 
During row-wise traversals, each cell should be mapped to multiple row headers, whereas column-wise traversals inherently incorporate header information into each column, facilitating more effective reasoning \cite{zhao-etal-2023-parser}.

\begin{figure}[t]
\centering
\input{fig/shot_num}
% \vspace{-0.5em}
\caption{
    % 我们的方法在WikiTQ上随着示例数量变化的性能
    Performance of \ours on WikiTQ with varying numbers of demonstrations.
}
\label{fig:shot_num}
\end{figure}
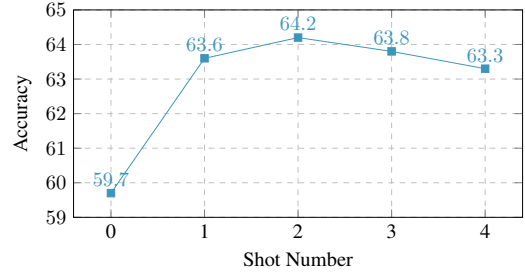

% 示例数量对我们方法的影响
\subsubsection{How does the number of demonstrations affect \ours?}
% 为了探索示例数量对我们方法的影响，我们在WikiTQ上使用Llama3.1-8B进行实验，结果如图所示
To investigate the effect of the number of demonstrations on \ours, we conduct experiments on WikiTQ using Llama3.1-8B, as illustrated in Figure~\ref{fig:shot_num}. 
% 具体地，我们的示例均选取于WikiTQ的训练集
All demonstrations were sampled from the WikiTQ training set.
% 我们发现
We observe that:
% 1. 示例数量从0变化到1时，性能上升最为明显
(\emph{i})~A substantial performance gain is observed when transitioning from zero-shot to one-shot prompting. 
% 因为示例有利于模型理解指令，并且可以根据示例中的推理过程模仿如何遍历表格得到答案，更好地激发了模型的表格推理能力
This suggests that a single demonstration significantly aids the model in comprehending the instruction and replicating the reasoning process for iterative row-wise table traversals, thus improving table reasoning capabilities.
% 2. 随着示例数量的增多，性能先上升后下降
(\emph{ii})~With a further increase in the number of demonstrations, performance initially improves but subsequently declines.
% 因为少量示例已经令模型很好地理解指令，学会逐行推理，更多的示例很难带来更多知识上的增益，反而容易限制模型的推理路径
A limited number of demonstrations is sufficient for the model to understand the instructions and learn the reasoning patterns. 
Additional demonstrations contribute little new information and may constrain the reasoning paths \cite{lin2024Alignment-icl,wan2025many-shot,zheng2025curse-cot-limitations-icl}.
% 因此，在主实验中我们选取了one-shot prompt，为了平衡了推理性能和效率
% Therefore, we employ the one-shot prompt in the main experiments to achieve a balance between inference performance and efficiency.
% 在主实验中我们选取了one-shot prompt，是因为其可以得到较好的性能，同时有着较高的推理效率
The one-shot prompt is chosen for our main experiments, balancing competitive performance with excellent inference efficiency.

%% file: tab/main.tex
\begin{tabular*}{.85\textwidth}{@{\extracolsep{\fill}}ll|lll}
\toprule
\textbf{Model} & \textbf{Method} & \textbf{WikiTQ} & \textbf{HiTab} & \textbf{TableBench}\\
\midrule
\multirow{2}{*}{Llama3.1-8B~\cite{dubey-etal-2024-llama3.1}} &  Short CoT & $57.9$ & $46.5$ & $31.5$\\
                           & \ours      & $63.6$ \green{($+2.7$)} & $\bm{56.6}$ \green{($+10.1$)} & $\bm{35.7}$ \green{($+4.2$)}\\
\cmidrule{1-5}
\multirow{2}{*}{R1\text{-}Llama-8B~\cite{guo2025deepseek-r1}} & Long CoT & $62.7$ & $49.7$ & $34.9$\\
                                        & \ours     & $\bm{63.7}$ \green{($+1.0$)} & $50.9$ \green{($+1.2$)} & $35.4$ \green{($+0.5$)}\\
\midrule
\multirow{2}{*}{Llama3.3-70B~\cite{dubey-etal-2024-llama3.1}} & Short CoT & $72.7$ & $66.9$ & $38.2$\\
                           & \ours      & $\bm{78.7}$ \green{($+6.0$)} & $\bm{72.4}$ \green{($+5.5$)} & $\bm{44.8}$ \green{($+6.6$)}\\
\cmidrule{1-5}
\multirow{2}{*}{R1\text{-}Llama-70B~\cite{guo2025deepseek-r1}} & Long CoT & $76.2$ & $67.4$ & $40.4$\\
                                      & \ours     & $78.3$ \green{($+2.1$)} & $68.6$ \green{($+1.2$)} & $42.8$ \green{($+2.4$)}\\
\midrule
\multirow{2}{*}{Qwen2.5-7B~\cite{yang2024qwen25}} & Short CoT & $52.2$ & $54.7$ & $30.9$\\
                          & \ours      & $\bm{61.7}$ \green{($+9.5$)} & $\bm{58.9}$ \green{($+4.2$)} & $34.9$ \green{($+4.0$)}\\
\cmidrule{1-5}
\multirow{2}{*}{R1\text{-}Qwen-7B~\cite{guo2025deepseek-r1}} & Long CoT & $53.3$ & $50.2$ & $34.2$\\
                                    & \ours     & $57.1$ \green{($+3.8$)} & $51.2$ \green{($+1.0$)} & $\bm{35.6}$ \green{($+1.4$)}\\
\midrule
\multirow{2}{*}{Qwen2.5-32B~\cite{yang2024qwen25}} & Short CoT & $69.2$ & $70.3$ & $35.9$\\
                          & \ours      & $75.6$ \green{($+6.4$)} & $\bm{76.6}$ \green{($+6.3$)} & $40.4$ \green{($+4.5$)}\\
\cmidrule{1-5}
\multirow{2}{*}{R1\text{-}Qwen-32B~\cite{guo2025deepseek-r1}} & Long CoT & $69.6$ & $70.8$ & $38.0$\\
                                      & \ours     & $\bm{76.9}$ \green{($+7.3$)} & $73.5$ \green{($+2.7$)} & $\bm{42.0}$ \green{($+4.0$)}\\
\bottomrule
\end{tabular*}

%% file: tab/SOTA_comparison.tex
\begin{tabular}{l|lc}
\toprule
\textbf{Dataset} & \textbf{Previous SOTA} & \textbf{\ours} \\
\midrule
WikiTQ & $78.0$~\cite{cao2025tablemaster} & \bm{$78.7$} \\
HiTab & \bm{$79.1$}~\cite{jiang2024sscot} & $76.7$ \\
TableBench & $43.9$~\cite{wu2024tablebench} & \bm{$44.8$} \\
\bottomrule
\end{tabular}

%% file: tab/ablation.tex
\begin{tabular*}{.75\textwidth}{@{\extracolsep{\fill}}lll|ccc}
\toprule
\textbf{Scale} & \textbf{Model} & \textbf{Method} & \textbf{WikiTQ} & \textbf{HiTab} & \textbf{TableBench}\\
\midrule
\multirow{6}{*}{8B} & \multirow{3}{*}{Llama3.1} &  \ours  & \bm{$63.6$} & $\bm{56.6}$ & $\bm{35.7}$\\
& & \textit{\quad w/o Iteration} &  $60.7$ & $55.5$ & $32.7$ \\
& & \textit{\quad w/o Traversal} &  $55.2$ & $42.2$ & $31.2$ \\
\cmidrule{2-6}
& \multirow{3}{*}{R1\text{-}Llama} & \ours     & $\bm{63.7}$ & \bm{$50.9$} & \bm{$35.4$}\\
& & \textit{\quad w/o Iteration} &  $56.6$ & $48.9$ & $31.5$ \\
& & \textit{\quad w/o Traversal} &  $46.8$ & $36.8$ & $25.7$ \\
\bottomrule
\end{tabular*}

%% file: fig/error.tex
% \resizebox{0.75\linewidth}{!}{
%     \begin{tikzpicture}[scale=0.6]
%         \small
%         \pie[
%             pos={8,0},
%             color={cred, cblue, cyellow, cwhite},
%             text=legend
%         ]{
%             36.0/Hallucination,
%             30.0/Overlooking,
%             20.0/Misunderstanding,
%             14.0/Linking
%         }
%     \end{tikzpicture}
% }

\resizebox{0.7\linewidth}{!}{
    \begin{tikzpicture}[scale=0.6]
        \small
        \pie[
            pos={8,0},
            color={spurple, sblue, sgreen},
            text=legend
        ]{
            66.0/Hallucination,
            20.0/Misunderstanding,
            14.0/Locating
        }
    \end{tikzpicture}
}

%% file: fig/multi_traversal.tex
\resizebox{0.7\linewidth}{!}{
    \begin{tikzpicture}[scale=0.6]
        \small
        \pie[
            pos={8,0},
            color={spurple, sblue},
            text=legend
        ]{
            68.3/Multi-Hop Reasoning,
            31.7/Reflection
            % 15.0/Summarization
        }
    \end{tikzpicture}
}

%% file: fig/shot_num.tex
\resizebox{.9\linewidth}{!}{
\begin{tikzpicture}
        \begin{axis}[
            width=10cm,
            height=5.5cm,
            ymin=59, ymax=65,
            xtick=data,
            xticklabels={0, 1, 2, 3, 4},
            ytick={59, 60, 61, 62, 63, 64, 65},
            bar width=20pt,
            xlabel={Shot Number},
            ylabel={Accuracy},
            nodes near coords,
            legend style={at={(0.5,1.05)},anchor=south,legend columns=-1},
            grid=both, % 添加此行以启用网格线
            grid style=dashed % 可选：设置网格线的样式为虚线
        ]
        % First line plot data
        \addplot[color=dblue, mark=square*] coordinates {
            (1, 59.7)
            (2, 63.6)
            (3, 64.2)
            (4, 63.8)
            (5, 63.3)
        };
        % \legend{\ourmethods, \ourmethodv}
        \end{axis}
    \end{tikzpicture}
}

%% file: tex/5.related.tex
\subsection{Table Reasoning}
% 表格推理任务旨在根据表格进行推理来完成用户的问题，被广泛用于金融、科研等数值密集的领域
The table reasoning task, which aims to answer user queries through inference over tabular data, is extensively applied in data-intensive domains such as finance and research \cite{jin2022table-survey,zhang2025table-survey}. 
% 目前，使用LLM完成表格推理任务成为了主流方法
Leveraging large language models (LLMs) has emerged as a prevalent method for table reasoning \cite{chen-2023-few-one,lu2025table-survey}. 
% 为了增强模型的表格推理能力，有的方法收集或者增广数据来微调模型
To enhance the table reasoning capability, researchers propose to collect or augment tabular data for fine-tuning \cite{zhang-etal-2024-tablellama,zhang2025tablellm,su2024tablegpt2}. 
% 但微调类工作耗费资源，而且容易降低模型的泛化性，所以前人提出了很多training-free的方法
However, the resource demands and potential reduction in generalization \cite{deng2025table-instruction-tuning} motivate training-free methods.
% 有的方法关注将问题分解来降低推理难度，比如TID首先从问题中抽取出三元组再转换为多个子问题，以完整地分解问题
Some methods focus on question decomposition to mitigate reasoning complexity \cite{ye2023dater,wu-feng-2024-protrix,jiang2024seek-solve}. 
For instance, TID~\cite{yang2025triples} extracts triples from the question and transforms them into sub-questions for comprehensive decomposition.
% 有的方法关注通过使用代码或工具来帮助模型推理，比如MACT结合a planning agent and a coding agent，为完成表格推理选择合适的动作和工具
Another direction involves the integration of programs or tools to facilitate reasoning \cite{jiang-etal-2023-structgpt,shi-etal-2024-ehragent,zhang-etal-2024-e5}, exemplified by MACT~\cite{zhou2025mact}, which employs a planning agent and a coding agent to select appropriate actions and tools for reasoning.

% 近些年，随着reasoning large language models (RLLMs)的发展，引入了Long CoT极大地提升了模型的推理能力，包括表格推理能力
Recent advancements in RLLMs demonstrate that the integration of Long CoT significantly improves their reasoning abilities, including table reasoning \cite{chen2025reasoning-survey,qian2025fino1}.
% 然而，Long CoT在表格推理任务上有着严重的表格内容幻觉的问题
% However, Long CoT suffers from High Cost and Low Reliability on the table reasoning task \cite{qin2024o1-replication,xu2025reasoning-survey-rl,zeng2025revisiting-test-truly}. 
However, Long Long CoT suffers from significant tabular content hallucination \cite{zeng2025revisiting-test-truly}.
% 所以，我们提出我们的方法，迭代的逐行遍历来推理，强制模型关注表格内容缓解了Long CoT的幻觉问题
% To address these limitations, we propose \ours, which is training-free and mitigates the hallucination issues associated with Long CoT by focusing on tabular content.
To address this, we propose an iteratively row-wise traversal method, which mitigates hallucination by forcing the model to focus on tabular content.

\subsection{Long CoT}
% reasoning large language models (RLLMs)，比如OpenAi O1和DeepSeek R1，通过引入Long CoT，scale test-time compute，显著提升了模型的推理能力，在很多复杂的下游推理任务上取得了一致的提升
RLLMs, such as OpenAI O1~\cite{openai2024openai-o1-card} and DeepSeek R1~\cite{guo2025deepseek-r1}, significantly improve reasoning capabilities by incorporating Long CoT with scaling reasoning length and iterative exploration and reflection, leading to consistent performance gains across diverse tasks \cite{snell2024scaling-test-time,aggarwal2025l1}. 
% 研究者一般通过有监督微调或者RL从base LLM训练获得RLLM
RLLMs are typically derived from base LLMs through supervised fine-tuning (SFT) or reinforcement learning (RL) \cite{chen2025reasoning-survey,chu2025sftmemorizes}. 
% 其中，SFT被用于从人类标注的数据或蒸馏数据中imitate advanced reasoning patterns
SFT aims to replicate sophisticated reasoning patterns from human-annotated or distilled data \cite{trung-etal-2024-reft,wen2025lightr1}. 
% 比如，s1和LIMO通过收集高质量、有难度且多样的问题，并精心标注rationale，分别获得了1000和817条训练数据，在其上SFT成功提升了模型的推理能力
For instance, s1~\cite{muennighoff2025s1} and LIMO~\cite{ye2025limo} enhance their reasoning abilities through SFT by collecting $1,000$ and $817$ high-quality training instances with meticulously labeled rationales, respectively.
% 而RL则通过自我学习和偏好优化来进一步提升模型的推理能力
RL further refines reasoning abilities through self-learning and preference optimization \cite{liu-etal-2024-aligning-survey,xu2025reasoning-survey-rl}. 
% 比如，zhang提出Process-based Self-Rewarding paradigm，在SFT初始化后，通过合成Step-wise Preference Data迭代DPO微调模型
For example, \citet{zhang2025process-self-rewarding} proposes a Process-based Self-Rewarding paradigm, which fine-tunes models using synthesized step-wise preference data.
% through Direct Preference Optimization. 

% 然而，前人工作都需要高质量数据，且耗费资源
However, previous works require high-quality data and exhibit significantly high cost
\cite{jiang2024slow1,qin2024o1-replication}. 
% 所以我们提出我们的方法，不需微调就能推理时间scale，并且超过推理模型的表格推理性能
Given that table reasoning tasks involve structured evidence, we propose \ours that enhances the table reasoning capabilities of non-reasoning LLMs without training.

%% file: tex/6.appendix.tex
\begin{table*}[t]
    \centering
    \small
    \input{tab/prompt}
    \caption{ 
        The demonstrations of \ours used on three datasets.
    }
    \label{tab:prompt}
\end{table*}

\section{Prompts}
\label{app:prompts}

\subsection{Demonstrations of \ours}
\label{subapp:Demonstrations of ours}
% 我们方法的prompt的Intruction如图所示，我们在本节展示在3个数据集上使用的示例，如表所示
The instructions for \ours are shown in Figure~\ref{fig:method}, so in this section, we present demonstrations used across three datasets in Table~\ref{tab:prompt}. 
% 我们在WikiTQ和TableBench上选用同一个来自WikiTQ训练集的示例，因为这两个数据集中的表格都是flat的，我们主要期望通过例子令模型理解逐行推理的过程
We select the same demonstration from the WikiTQ training set for both WikiTQ and TableBench, as the tables in these two datasets are flat. 
Our primary goal is to help the model understand the process of row-wise table traversals through the demonstration. 
% 而HiTab中的表格为嵌套的，由于其特殊性，我们选取了HiTab训练集中的例子，以便令模型更好地理解这种表格结构
In contrast, the tables in HiTab are hierarchical. 
Due to this distinct structure, we select the demonstration from the HiTab training set to better facilitate the understanding of the table structure.

\begin{table*}[t]
    \centering
    \small
    \input{tab/prompt_ablation}
    \caption{ 
        The prompts used in the ablation experiments.
    }
    \label{tab:prompt_ablation}
\end{table*}

\subsection{Prompts for ablation experiments}
\label{subapp:Prompts for ablation experiments}
We show the prompts used in ablation experiments in Table~\ref{tab:prompt_ablation}.
% 消融实验使用的示例和主实验保持一致，只是将示例的推理对应地移除迭代过程和遍历过程
In the ablation study, the demonstrations used are consistent with those in the main experiments, with the corresponding iterative and traversal processes removed from the reasoning process.

\section{Additional Experiments}
\label{app:additional experiments}

\begin{table}[t]
    \centering
    \small
    \input{tab/Long_CoT_few_shot}
    \caption{ 
        % 我们的方法和基线的性能对比
        Performance of Long CoT using R1-Llama-8B with zero-shot and few-shot.
    }
    \label{tab:Long_CoT_few_shot}
\end{table}

\subsection{Long CoT with few-shot prompt}
\label{subapp:Long CoT with few-shot prompt}
% 在本小节，我们展示使用few-shot prompt的Long CoT的性能，使用R1-Llama-8B时，如表所示
In this subsection, we present the performance of Long CoT using few-shot prompts with R1-Llama-8B, as shown in Table~\ref{tab:Long_CoT_few_shot}. 
% 可以发现，在三个数据集上，Long CoT的性能相比zero-shot都大幅下降，因此在主实验中我们采用了zero-shot的prompt
It can be observed that, across three datasets, the performance of Long CoT significantly declines compared to the zero-shot setting. 
Therefore, in the main experiments, we employ zero-shot prompts.

\begin{table}[t]
    \centering
    \small
    \input{tab/WikiTQ_comparison}
    \caption{ 
        % 我们的方法和基线的性能对比
        Performance comparison between \ours and previous methods with comparable scale models on WikiTQ.
    }
    \label{tab:WikiTQ_comparison}
\end{table}

\begin{table}[t]
    \centering
    \small
    \input{tab/HiTab_comparison}
    \caption{ 
        % 我们的方法和基线的性能对比
        Performance comparison between \ours and previous methods with comparable scale models on HiTab.
    }
    \label{tab:HiTab_comparison}
\end{table}

\begin{table}[t]
    \centering
    \small
    \input{tab/TableBench_comparison}
    \caption{ 
        % 我们的方法和基线的性能对比
        Performance comparison between \ours and previous methods with comparable scale models on TableBench.
    }
    \label{tab:TableBench_comparison}
\end{table}

\subsection{Comparison with previous methods}
\label{subapp:Comparison with previous methods}
% 在本小节，我们展示我们的方法和更多前人工作的比较，如表4、5和6所示
In this subsection, we present a comparison of \ours with previous works, as shown in Table~\ref{tab:WikiTQ_comparison}, Table~\ref{tab:HiTab_comparison}, and Table~\ref{tab:TableBench_comparison}. 
% 我们的方法在WikiTQ和TableBench上都达到了SOTA，在HiTab上和前人方法可比，证明了我们方法的有效性
\ours achieves state-of-the-art performance on WikiTQ and TableBench, and performs comparably to prior methods on HiTab, demonstrating its effectiveness.
% 我们的方法优于前人方法，因为\ours通过细节、迭代的探索和反思优化了表格推理过程
\ours surpasses prior methods by optimizing the table reasoning process through detailed, iterative exploration and reflection.

% 值得一提的是，Table-Critic提出了一个用于表格推理的多agent系统，包括Judge用于识别错误，Critic对识别到的错误进行分析，一个Refiner用于修改错误，和一个Curator用于收集critic知识，以提进一步提升批评的质量
Notably, Table-Critic~\cite{yu2025table-critic} introduces a multi-agent system for table reasoning, comprising a Judge to identify errors, a Critic to analyze these identified errors, a Refiner to rectify them, and a Curator to aggregate critic knowledge for enhanced critique quality.
% \ours超越了Table-Critic，说明了\ours不仅可以对已有的推理进行有效反思，逐行遍历的推理模式通过细粒度的推理进一步提升了表格推理能力
\ours surpasses Table-Critic using the same LLM, demonstrating not only effective reflection on previous reasoning but also sequential scaling through row-wise traversal, leading to improved table reasoning capabilities.

\begin{figure}[t]
\centering
\input{fig/error_RLLM}
% \vspace{-0.5em}
\caption{
    % Long CoT因为呈现出的错误类型而不如\ours
    Long CoT underperforms \ours with RLLMs due to the error types, with their distribution.
}
\label{fig:error_RLLM}
\end{figure}
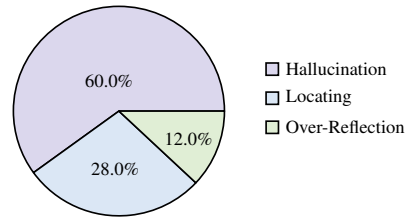

\subsection{Why \ours with RLLMs outperforms Long CoT?}
\label{subapp:ours with RLLMs outperforms Long CoT}
% 为了更具体地分析为什么我们方法 with RLLMs的性能超过Long CoT，我们随机选取了我们的方法使用R1-Llama-8B在WikiTQ上的做对，但Long CoT使用R1-Llama-8B没做对的instance，手动分析了原因，比例如图所示
To analyze specifically why \ours with RLLMs outperforms Long CoT, we randomly select instances from WikiTQ where \ours using R1-Llama-8B provided the correct answer, but Long CoT using R1-Llama-8B failed. 
We manually analyze the reasons for these discrepancies, with the distribution shown in Figure~\ref{fig:error_RLLM}. 
% 其中，幻觉和Locating如3.2.2节的介绍
Among them, Hallucination and Locating are as described in \S\ref{subsec:ours Outperforms Long CoT}. 
% Over-Reflection指的是模型因为反思将原本正确的答案改错，或反思次数过多超出模型限制导致没有得到最终答案
Over-Reflection refers to cases where the reflection process led to an originally correct answer being changed to incorrect, or where excessive reflections exceeding the token limits prevented a final answer from being generated. 
% 可以发现，\ours很大程度上缓解了Long CoT的幻觉
The results indicate that:
(\emph{i})~\ours significantly mitigates the hallucination issue prevalent in Long CoT.
% 并且，通过顺序的逐行遍历，定位到所有相关信息的能力增强
(\emph{ii})~The sequential row-wise traversal enhances the ability to locate all relevant information. 
% 而且，\ours可以一定程度上缓解Over-Reflection，因为\ours规定了模型遍历表格进行反思，缓解了无效或错误的反思
(\emph{iii})~\ours can alleviate Over-Reflection to some extent by guiding the reflection process through structured table traversal, thus reducing ineffective or erroneous reflections.

\section{Case Study}
\label{app:case study}

\begin{figure*}[t]
    \centering
    \includegraphics[width=0.95\linewidth]{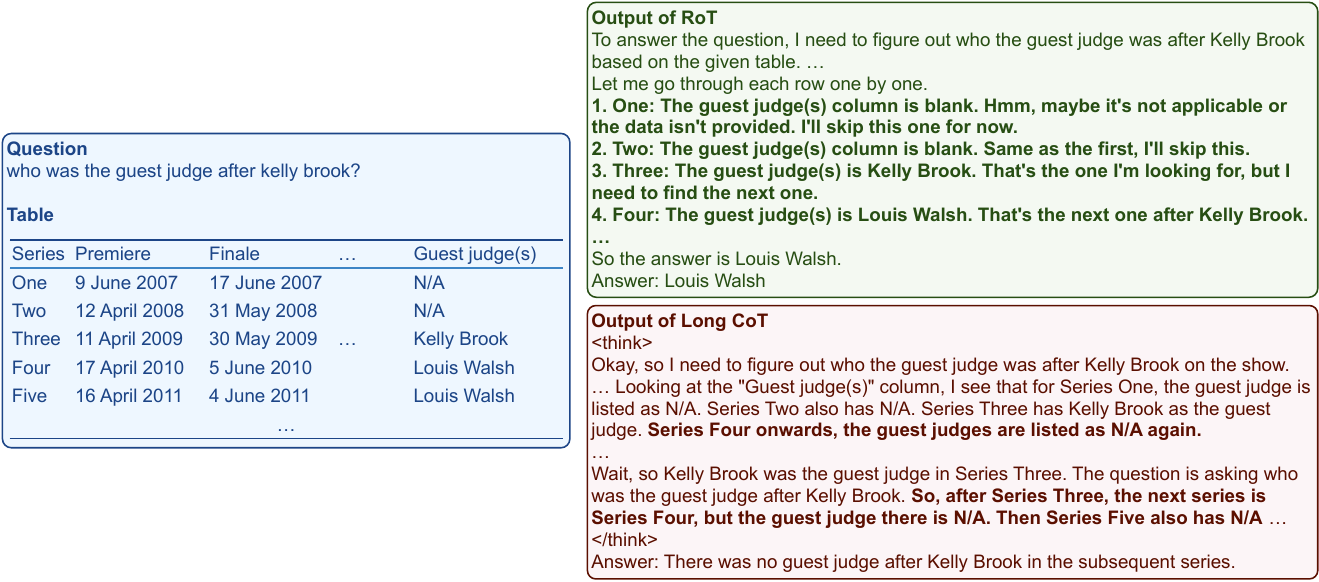}
    % \vspace{-1.5em}
    \caption{
    % 我们的方法因为Multi-Step Reasoning而迭代row-wise table traversals的例子
    \ours outperforms Long CoT due to the hallucination issue of Long CoT.
    % row-wise table traversal process被高亮蓝色，幻觉的推理步骤被高亮红色
    The row-wise table traversal process is highlighted in \green{\textbf{green}}, and the hallucinated reasoning steps are highlighted in \red{\textbf{red}}.
    }
    \label{fig:case_hallucination}
\end{figure*}

% \begin{figure*}[t]
%     \centering
%     \includegraphics[width=0.95\linewidth]{fig/case_overlooking.pdf}
%     % \vspace{-1.5em}
%     \caption{
%     % 我们的方法因为Multi-Step Reasoning而迭代row-wise table traversals的例子
%     \ours outperforms Long CoT due to the overlooking issue of Long CoT.
%     % row-wise table traversal process被高亮蓝色，幻觉的推理步骤被高亮红色
%     The row-wise table traversal process is highlighted in blue, and the incorrect reasoning steps with overlooking are highlighted in red.
%     }
%     \label{fig:case_overlooking}
% \end{figure*}

\begin{figure*}[t]
    \centering
    \includegraphics[width=0.95\linewidth]{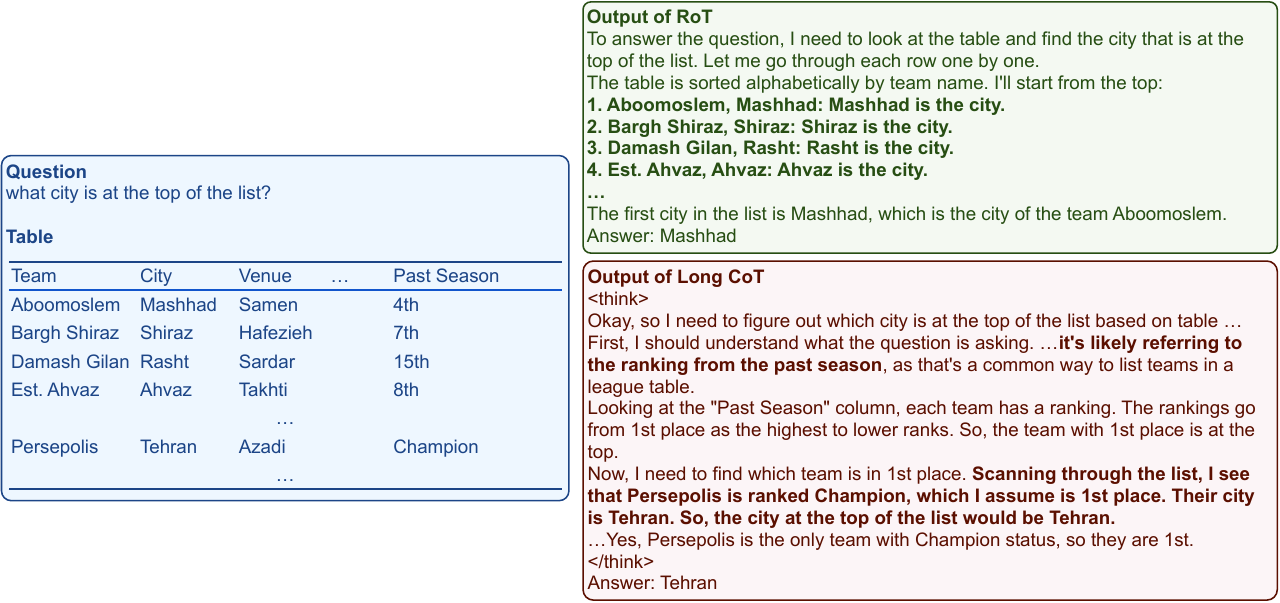}
    % \vspace{-1.5em}
    \caption{
    % 我们的方法因为Multi-Step Reasoning而迭代row-wise table traversals的例子
    \ours outperforms Long CoT due to the misunderstanding issue of Long CoT.
    % row-wise table traversal process被高亮蓝色，幻觉的推理步骤被高亮红色
    The row-wise table traversal process is highlighted in \green{\textbf{green}}, and the misunderstanding reasoning steps are highlighted in \red{\textbf{red}}.
    }
    \label{fig:case_misunderstanding}
\end{figure*}

\begin{figure*}[t]
    \centering
    \includegraphics[width=0.95\linewidth]{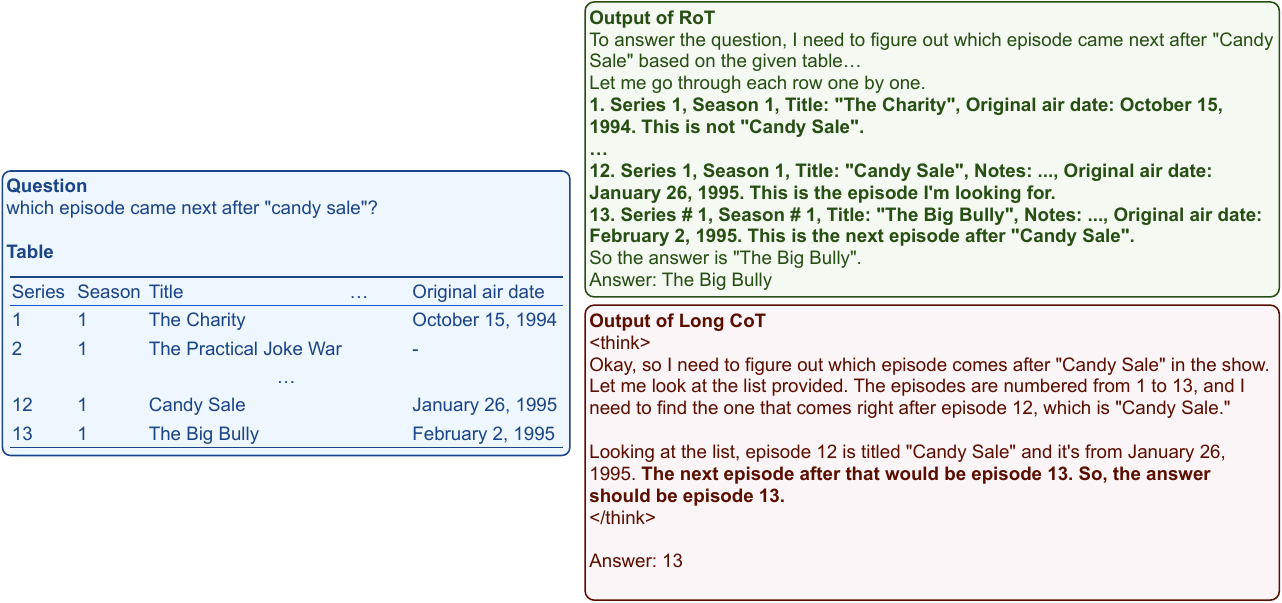}
    % \vspace{-1.5em}
    \caption{
    % 我们的方法因为Multi-Step Reasoning而迭代row-wise table traversals的例子
    \ours outperforms Long CoT due to the locating error of Long CoT.
    % row-wise table traversal process被高亮蓝色，错误链接的推理步骤被高亮红色
    The row-wise table traversal process is highlighted in \green{\textbf{green}}, and the incorrect linking reasoning steps are highlighted in \red{\textbf{red}}.
    }
    \label{fig:case_linking}
\end{figure*}

\subsection{Case study of \ours compared with Long CoT}
\label{subapp:Case study of ours compared with Long CoT}

% 我们展示了因为不同原因\ours打败Long CoT的例子，如图所示
We present examples where \ours outperforms Long CoT for distinct reasons, as illustrated in Figure~\ref{fig:case_hallucination}, Figure~\ref{fig:case_misunderstanding}, and Figure~\ref{fig:case_linking}.

\begin{figure*}[t]
    \centering
    \includegraphics[width=0.95\linewidth]{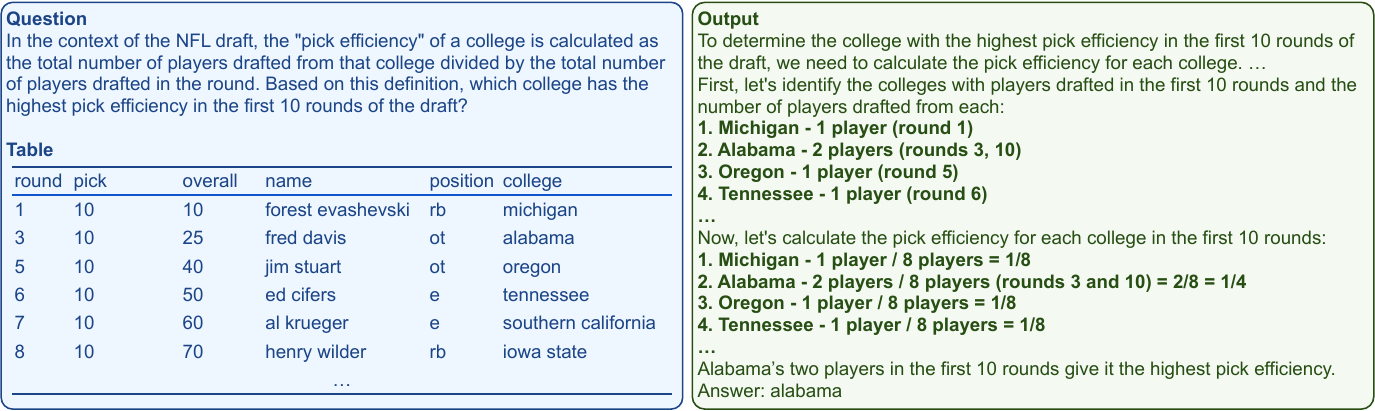}
    % \vspace{-1.5em}
    \caption{
    % 我们的方法因为Multi-Step Reasoning而迭代row-wise table traversals的例子
    \ours performs iterative row-wise table traversals due to Multi-Hop Reasoning.
    The row-wise table traversal process is highlighted in \green{\textbf{green}}.
    }
    \label{fig:traversals_multi_step}
\end{figure*}

\begin{figure*}[t]
    \centering
    \includegraphics[width=0.95\linewidth]{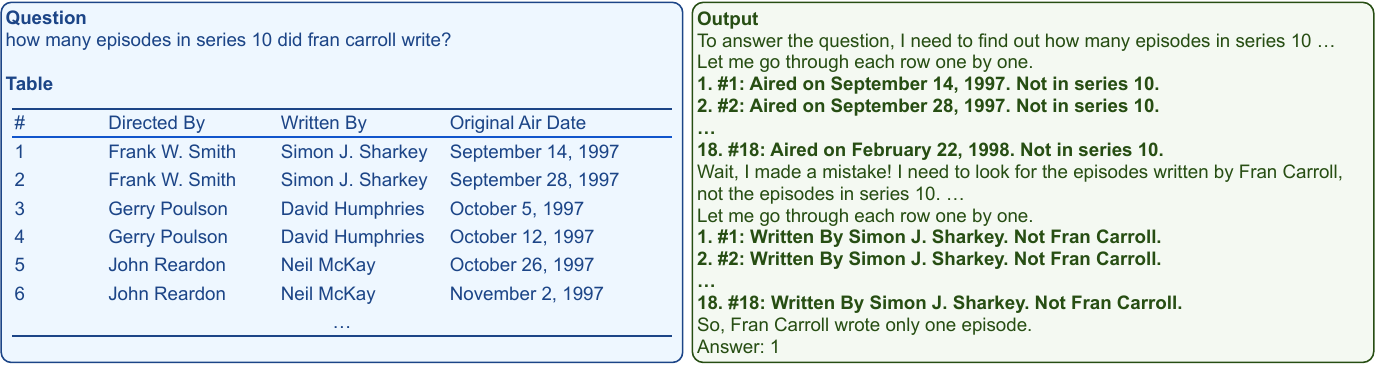}
    % \vspace{-1.5em}
    \caption{
    % 我们的方法因为Multi-Step Reasoning而迭代row-wise table traversals的例子
    \ours performs iterative row-wise table traversals due to Reflection.
    The row-wise table traversal process is highlighted in \green{\textbf{green}}.
    }
    \label{fig:traversals_reflection}
\end{figure*}

% \begin{figure*}[t]
%     \centering
%     \includegraphics[width=0.95\linewidth]{fig/traversals_summary.pdf}
%     % \vspace{-1.5em}
%     \caption{
%     % 我们的方法因为Multi-Step Reasoning而迭代row-wise table traversals的例子
%     \ours performs iterative row-wise table traversals due to Summary.
%     The row-wise table traversal process is highlighted.
%     }
%     \label{fig:traversals_summary}
% \end{figure*}

\subsection{Case study of \ours with iterative traversals}
\label{subapp:Case study of ours with iterative traversals}
% 我们分别展示了\ours performs iterative row-wise table traversals对应的3种原因的例子，在图5、6和7中
We present examples of the three reasons for \ours performing iterative row-wise table traversals in Figure~\ref{fig:traversals_multi_step} and Figure~\ref{fig:traversals_reflection}.

%% file: tab/prompt.tex
\begin{tabular}{p{0.9\textwidth}}
\toprule
\textbf{The demonstration of \ours used on WikiTQ and TableBench} \\
\midrule
% Your task is to think step by step by traversing the given table to solve the question. \\
% Note that: \\
% 1. You must traverse the whole table correctly and extract the relevant information.\\
% 2. Represent your answer with: Answer: <Your Answer>.\\
% Here is an example:\\
% \\
% ---\\
% \\
| Parish | Locality | Parish Priest | Founded | Closed |\\
|:---|:---|:---|:---|:---|\\
| St Mary | Bacup | Fr Frank Thorpe | 1852 | ---- |\\
| Our Immaculate Mother \& St Anselm | Whitworth | Fr Frank Thorpe | 1860 | ---- |\\
| St Joseph | Stacksteads | ---- | 1947 | 2005 |\\
| St Joseph \& St Peter | Newchurch-In-Rossendale | Fr Philip Boast | 1915 | ---- |\\
| The Immaculate Conception | Haslingden | Fr Canon John Mackie | 1854 | ---- |\\
| St Veronica (Chapel of Ease) | Helmshore | Served from The Immaculate Conception | 1959 | ---- |\\
| St James the Less | Rawtenstall | Fr David Lupton, Rural Dean | 1828 | ---- |\\
Question:\\
what's the number of parishes founded in the 1800s?\\
Solution:\\
To answer the question, I need to figure out how many parishes were founded in the 1800s based on the given table. Let me look at the table again. The table has several columns: Parish, Locality, Parish Priest, Founded, and Closed. My task is to count how many parishes were established in the 1800s.\\
\\
First, I should understand what the "Founded" column represents. It likely indicates the year the parish was established. So, I need to look at each row and check the year under the "Founded" column. If the year is in the 1800s, I'll count that parish.\\
\\
Let me go through each row one by one.\\
\\
1. St Mary, Bacup: Founded in 1852. That's in the 1800s, so that's one.\\
2. Our Immaculate Mother \& St Anselm, Whitworth: Founded in 1860. Also in the 1800s, so that's two.\\
3. St Joseph, Stacksteads: The "Founded" column is blank. Hmm, maybe it's not applicable or the data isn't provided. I'll skip this one for now.\\
4. St Joseph \& St Peter, Newchurch-In-Rossendale: Founded in 1915. That's in the early 1900s, so not in the 1800s.\\
5. The Immaculate Conception, Haslingden: Founded in 1854. That's another one, so three.\\
6. St Veronica (Chapel of Ease), Helmshore: Founded in 1959. That's in the 1900s, so not relevant.\\
7. St James the Less, Rawtenstall: Founded in 1828. That's in the 1800s, so four.\\
\\
So the answer is four.\\
\\
Answer: 4\\
% \\
% ---\\
% \\
% Based on the above example, you need to traverse the table below and answer the question.\\
% \\
% \{Table\}\\
% Question:\\
% \{Question\}\\
% Solution:\\
\bottomrule
\end{tabular}

\begin{tabular}{p{0.9\textwidth}}
\toprule
\textbf{The demonstration of \ours used on HiTab} \\
\midrule
admissions to adult custody, by sex and by indigenous identity and jurisdiction, 2018/2019\\
| jurisdiction | male | male | male | male | male | male | female | female | female | female | female | female |\\
|:---|:---|:---|:---|:---|:---|:---|:---|:---|:---|:---|:---|:---|\\
| jurisdiction | indigenous | indigenous | non-indigenous | non-indigenous | unknown indigenous identity | unknown indigenous identity | indigenous | indigenous | non-indigenous | non-indigenous | unknown indigenous identity | unknown indigenous identity |\\
| jurisdiction | number | percent change in number from 2017/2018 | number | percent change in number from 2017/2018 | number | percent change in number from 2017/2018 | number | percent change in number from 2017/2018 | number | percent change in number from 2017/2018 | number | percent change in number from 2017/2018 |\\
| newfoundland and labrador | 333 | -7 | 1,214 | -9 | 0 | .. | 39 | -11 | 187 | -4 | 0 | .. |\\
| prince edward island | 25 | 9 | 606 | 11 | 4 | -33 | 4 | -56 | 129 | 39 | 1 | 0 |\\
| nova scotia | 337 | 6 | 3,676 | 3 | 18 | 200 | 108 | 21 | 512 | -8 | 8 | 14 |\\
| new brunswick | 362 | -10 | 4,319 | 4 | 3 | 0 | 148 | -3 | 916 | 17 | 0 | -100 |\\
...\\
% | quebec | 2,080 | 9 | 35,887 | -6 | 0 | .. | 450 | 11 | 4,677 | -4 | 0 | .. |
% | ontario | 6,441 | -6 | 49,550 | -7 | 0 | .. | 1,615 | -1 | 7,212 | -5 | 0 | .. |
% | manitoba | 16,126 | -6 | 6,051 | -7 | 0 | .. | 4,919 | -6 | 1,044 | 3 | 0 | .. |
% | saskatchewan | 7,917 | -1 | 2,711 | 2 | 237 | -29 | 1,922 | 1 | 349 | -12 | 16 | -11 |
% | alberta | 12,235 | -13 | 18,358 | -16 | 385 | -15 | 3,535 | -18 | 2,875 | -14 | 96 | -28 |
% | british columbia | 6,600 | -8 | 14,448 | -10 | 75 | -3 | 1,016 | -18 | 1,282 | -12 | 10 | 25 |
% | yukon | 254 | -6 | 129 | -22 | 0 | .. | 27 | 0 | 11 | -21 | 0 | .. |
% | northwest territories | 856 | 1 | 97 | -21 | 9 | -55 | 62 | -16 | 6 | -25 | 1 | -50 |
% | nunavut | 1,325 | 51 | 0 | . | 72 | 80 | 74 | 14 | 0 | . | 4 | 100 |
% | total-provinces and territories | 54,891 | -6 | 137,046 | -7 | 803 | -14 | 13,919 | -8 | 19,200 | -6 | 136 | -21 |
% | federal | 1,994 | 6 | 5,001 | 2 | 0 | .. | 233 | 4 | 330 | -3 | 0 | .. |
Question:\\
what is the percentage of admissions of non-indigenous males decreased in the provinces and territories compared with the previous year?\\
Solution:\\
Okay, so I need to figure out the percentage decrease in admissions of non-indigenous males across the provinces and territories compared to the previous year. Let me look at the table provided.\\
% \\
% First, I see that the table has several columns. The headers are a bit confusing at first glance, but I think the columns are organized by jurisdiction, then for males, there are categories like indigenous, non-indigenous, and unknown, each with their numbers and percent changes. The same goes for females, but I'm only concerned with males here.\\
% \\
% Looking at the structure, each jurisdiction has multiple entries. For example, Newfoundland and Labrador has numbers under male indigenous, male non-indigenous, and male unknown. Each of these has a number and a percent change from the previous year. I'm specifically interested in the non-indigenous males.\\
% \\
I need to find the percentage change for non-indigenous males in each province and territory. Let me go through each row:\\
% \\
1. Newfoundland and Labrador: Non-indigenous males have a percent change of -9\%.\\
2. Prince Edward Island: Non-indigenous males have a percent change of 11\%.\\
3. Nova Scotia: Non-indigenous males have a percent change of 3\%...\\
% 4. New Brunswick: Non-indigenous males have a percent change of 4\%.\\
% ...\\
% 5. Quebec: Non-indigenous males have a percent change of -6%.
% 6. Ontario: Non-indigenous males have a percent change of -7%.
% 7. Manitoba: Non-indigenous males have a percent change of -7%.
% 8. Saskatchewan: Non-indigenous males have a percent change of 2%.
% 9. Alberta: Non-indigenous males have a percent change of -16%.
% 10. British Columbia: Non-indigenous males have a percent change of -10%.
% 11. Yukon: Non-indigenous males have a percent change of -22%.
% 12. Northwest Territories: Non-indigenous males have a percent change of -21%.
% 13. Nunavut: Non-indigenous males have a percent change of . (which I think means no change or data not available, so maybe 0% or not considered).
% 14. Total-provinces and territories: Non-indigenous males have a percent change of -7%.
% 15. Federal: Non-indigenous males have a percent change of 2%.
It shows a -7\% change for non-indigenous males. That seems to be the overall percentage decrease across all provinces and territories combined. I think that's the answer because the total row aggregates all the data, so it's the most accurate representation of the overall percentage change.\\
% The percentage decrease in admissions of non-indigenous males across all provinces and territories compared to the previous year is 7\%.\\
\\
Answer: 7\\
\bottomrule
\end{tabular}

%% file: tab/prompt_ablation.tex
\begin{tabular}{p{0.9\textwidth}}
\toprule
\textbf{The prompt of \ours without Iteration} \\
\midrule
Your task is to think step by step by traversing the given table to solve the question. \\
Note that: \\
1. You must traverse the table row by row once and only once.\\
2. Represent your answer with: Answer: <Your Answer>.\\
Here is an example:\\
...\\
Based on the above example, you need to traverse the table below and answer the question.\\
\\
\{Table\}\\
Question:\\
\{Question\}\\
Solution:\\
\bottomrule
\end{tabular}

\begin{tabular}{p{0.9\textwidth}}
\toprule
\textbf{The prompt of \ours without Traversal} \\
\midrule
Your task is to think step by step to solve the question. \\
Note that: \\
1. You must reflect your reasoning iteratively until you get a correct answer.\\
2. Represent your answer with: Answer: <Your Answer>.\\
Here is an example:\\
...\\
Based on the above example, you need to answer the question below.\\
\\
\{Table\}\\
Question:\\
\{Question\}\\
Solution:\\
\bottomrule
\end{tabular}

%% file: tab/Long_CoT_few_shot.tex
\begin{tabular}{ll|cccc}
\toprule
\textbf{Dataset} & \textbf{Method} & \textbf{Perfromance} \\
\midrule
\multirow{2}{*}{WikiTQ} & Long CoT (zero-shot) & \bm{$62.7$} \\
 & Long CoT (one-shot) & $45.1$ \\
\midrule
\multirow{2}{*}{HiTab} & Long CoT (zero-shot) & \bm{$49.7$} \\
 &Long CoT (one-shot) & $35.4$ \\
\midrule
\multirow{2}{*}{TableBench} & Long CoT (zero-shot) & \bm{$34.9$} \\
 &Long CoT (one-shot) & $25.3$ \\
\bottomrule
\end{tabular}

%% file: tab/WikiTQ_comparison.tex
% \begin{tabular}{l|lc}
% \toprule
% \textbf{Method} & \textbf{Model} & \textbf{Accuracy} \\
% \midrule
% Binder~\cite{cheng2023binder} & Llama3.3-70B & $52.2$ \\
% Dater~\cite{ye2023dater} & Llama3.3-70B & $59.5$ \\
% Chain-of-Table~\cite{wang2024chainoftable} & Llama3.3-70B & $62.1$ \\
% PoTable~\cite{mao2025potable} & Llama3.1-70B & $65.6$ \\
% FlexTaF~\cite{zhang2024flextaf} & Llama3-70B & $69.9$ \\
% Table-Critic~\cite{yu2025table-critic} & Llama3.3-70B & $70.1$ \\
% MACT~\cite{zhou2025mact} & Qwen2-72B & $72.6$ \\
% SS-CoT~\cite{jiang2024sscot} & Llama3.1-70B & $76.8$ \\
% TableMaster~\cite{cao2025tablemaster} & Llama3.1-70B & $78.0$ \\
% \midrule
% \ours & Llama3.3-70B & \bm{$78.7$} \\
% \bottomrule
% \end{tabular}

% \begin{tabular}{l|lc}
% \toprule
% \textbf{Method} & \textbf{Model} & \textbf{Accuracy} \\
% \midrule
% Binder & Llama3.3-70B & $52.2$ \\
% Dater & Llama3.3-70B & $59.5$ \\
% Chain-of-Table & Llama3.3-70B & $62.1$ \\
% PoTable & Llama3.1-70B & $65.6$ \\
% FlexTaF & Llama3-70B & $69.9$ \\
% Table-Critic & Llama3.3-70B & $70.1$ \\
% MACT & Qwen2-72B & $72.6$ \\
% % TIDE & GPT-3.5 & $75.0$ \\
% SS-CoT & Llama3.1-70B & $76.8$ \\
% TableMaster & Llama3.1-70B & $78.0$ \\
% \midrule
% \ours & Llama3.3-70B & \bm{$78.7$} \\
% \bottomrule
% \end{tabular}

\begin{tabular}{l|c}
\toprule
\textbf{Method} & \textbf{Accuracy} \\
\midrule
Llama3-70B & \\
FlexTaF~\cite{zhang2024flextaf} & $69.9$ \\
\midrule
Llama3.1-70B & \\
PoTable~\cite{mao2025potable} & $65.6$ \\
SS-CoT~\cite{jiang2024sscot} & $76.8$ \\
TableMaster~\cite{cao2025tablemaster} & $78.0$ \\
\midrule
Qwen2-72B & \\
MACT~\cite{zhou2025mact}  & $72.6$ \\
\midrule
Llama3.3-70B & \\
Binder~\cite{cheng2023binder} & $52.2$ \\
Dater~\cite{ye2023dater} & $59.5$ \\
Chain-of-Table~\cite{wang2024chainoftable} & $62.1$ \\
Table-Critic~\cite{yu2025table-critic} & $70.1$ \\
% TIDE & GPT-3.5 & $75.0$ \\
% \midrule
\ours & \bm{$78.7$} \\
\bottomrule
\end{tabular}

%% file: tab/HiTab_comparison.tex
% \begin{tabular}{l|lc}
% \toprule
% \textbf{Method} & \textbf{Model} & \textbf{Accuracy} \\
% \midrule
% Zhao et al. & GPT-3.5 & $50.0$ \\
% Cao et al. & code-davinci-002 & $69.3$ \\
% GraphOTTER & Qwen2-72B & $72.7$ \\
% SS-CoT & Llama3.1-70B & \bm{$79.1$} \\
% \midrule
% \ours & Qwen2.5-32B & $76.6$ \\
% \bottomrule
% \end{tabular}

\begin{tabular}{l|c}
\toprule
\textbf{Method} & \textbf{Accuracy} \\
\midrule
GPT-3.5 & \\
\citet{zhao-etal-2023-parser} & $50.0$ \\
\midrule
code-davinci-002 & \\
\citet{cao-etal-2023-api} & $69.3$ \\
\midrule
Qwen2-72B & \\
GraphOTTER~\cite{li-etal-2025-graphotter} & $72.7$ \\
\midrule
Llama3.1-70B & \\
SS-CoT~\cite{jiang2024sscot} & \bm{$79.1$} \\
\midrule
Qwen2.5-32B & \\
\ours & $76.6$ \\
\bottomrule
\end{tabular}

%% file: tab/TableBench_comparison.tex
\begin{tabular}{l|c}
\toprule
\textbf{Method} & \textbf{Accuracy} \\
\midrule
Llama3.1-70B & \\
\citet{wu2024tablebench} & $43.9$ \\
\midrule
Llama3.3-70B & \\
\ours & \bm{$44.8$} \\
\bottomrule
\end{tabular}

%% file: fig/error_RLLM.tex
\resizebox{0.7\linewidth}{!}{
    \begin{tikzpicture}[scale=0.6]
        \small
        \pie[
            pos={8,0},
            color={spurple, sblue, sgreen},
            text=legend
        ]{
            60.0/Hallucination,
            28.0/Locating,
            12.0/Over-Reflection
        }
    \end{tikzpicture}
}